\documentclass[11pt]{article}

\usepackage[final]{acl}

\usepackage{times}
\usepackage{latexsym}
\usepackage[utf8]{inputenc}
\usepackage{times}
\usepackage{latexsym}
\usepackage{booktabs}
\usepackage[T1]{fontenc}
\usepackage[utf8]{inputenc}
\usepackage{microtype}
\usepackage{inconsolata}
\usepackage{graphicx}
\usepackage{marvosym}
\usepackage{hyperref}
\usepackage{algpseudocode}
\usepackage{xcolor,colortbl}
\usepackage{amssymb}
\usepackage{colortbl}
\definecolor{mygray}{gray}{.90}
\usepackage{multirow}
\usepackage{pifont}

\usepackage{amssymb}
\usepackage{mathtools}
\usepackage{amsthm}
\usepackage[capitalize,noabbrev]{cleveref}
\theoremstyle{plain}

\theoremstyle{definition}

\theoremstyle{remark}

\usepackage[textsize=tiny]{todonotes}

\usepackage[T1]{fontenc}

\usepackage[utf8]{inputenc}

\usepackage{microtype}

\usepackage{inconsolata}

\usepackage{graphicx}
\usepackage{amsmath} 
\usepackage{multirow}

\title{Think before Go: Hierarchical Reasoning for Image-goal Navigation}

\author{
Pengna Li$^{1,3}$\thanks{Co-first Authors.\\
\indent\{sauerfisch, wukangyi747600\}@stu.xjtu.edu.cn\\
\indent$^\dagger$ Project Leader.\\
\indent\textsuperscript{\Letter} Corresponding Author.}
\qquad Kangyi Wu$^{1,*}$
\qquad Shaoqing Xu$^{2,3,\dagger}$
\qquad Fang Li$^{2,3}$\\
\textbf{Lin Zhao$^{4}$}
\qquad \textbf{Long Chen$^{3}$}
\qquad \textbf{Zhixin Yang$^{2}$}
\qquad \textbf{Nanning Zheng$^{1}$\textsuperscript{\Letter}}\\[5pt]
$^1$National Key Laboratory of Human-Machine Hybrid Augmented Intelligence,\\
National Engineering Research Center for Visual Information and Applications,\\
Institute of Artificial Intelligence and Robotics, Xi'an Jiaotong University\\
$^2$University of Macau, $^3$Xiaomi EV\\
$^4$School of Automation, Beijing Institute of Technology
}

\begin{document}
\maketitle
\begin{abstract}
Image-goal navigation steers an agent to a target location specified by an image in unseen environments. 
Existing methods primarily handle this task by learning an end-to-end navigation policy, which compares the similarities of target and observation images and directly predicts the actions. 
However, when the target is distant or lies in another room, such methods fail to extract informative visual cues, leading the agent to wander around.
Motivated by the human cognitive principle that deliberate, high-level reasoning guides fast, reactive execution in complex tasks, we propose Hierarchical Reasoning Navigation~(HRNav), a framework that decomposes image-goal navigation into high-level planning and low-level execution. 
In high-level planning, a vision-language model is trained on a self-collected dataset to generate a short-horizon plan, such as whether the agent should walk through the door or down the hallway. This downgrades the difficulty of the long-horizon task, making it more amenable to the execution part. 
In low-level execution, an online reinforcement learning policy is utilized to decide actions conditioned on the short-horizon plan. We also devise a novel Wandering Suppression Penalty~(WSP) to further reduce the wandering problem. 
Together, these components form a hierarchical framework for Image-goal Navigation.
Extensive experiments in both simulation and real-world environments demonstrate the superiority of our method.
\end{abstract}

\section{Introduction}
Image-goal Navigation~\cite{zhu2017target} has emerged as a fundamental problem in embodied navigation~\cite{zheng2024towards,gan2020look}, which aims to enable an agent to autonomously move within an unseen environment to reach a target specified by an image. 
Motivated by its broad potential in last-mile delivery and household robots~\cite{wasserman2023last,majumdar2022zson,krantz2023navigating}, this task has received growing attention from the research community.

\begin{figure}[t]
  \centering
  \includegraphics[width=\linewidth]{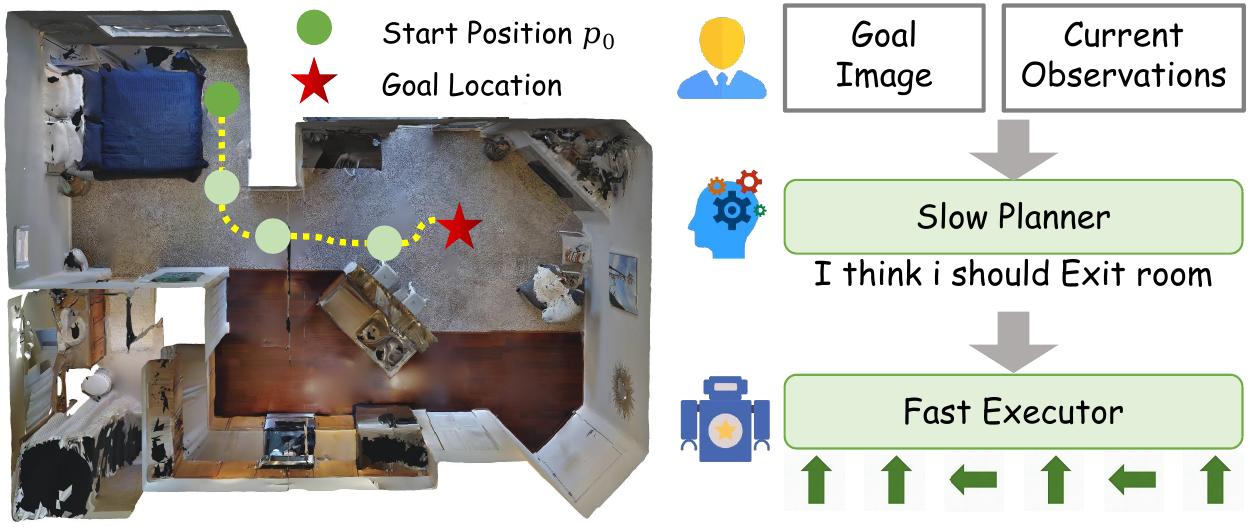}
  \caption{\textbf{Our proposed HRNav.}
  We adopt a two-level hierarchy: given goal image and current observations, a \emph{slow} planner predicts short-horizon plans, while a \emph{fast} executor follows these short-horizon plans to produce low-level actions, enabling more efficient navigation.}
  \label{fig:intro_hierarchy}
\end{figure}

Despite the rapid growth, image-goal navigation remains highly challenging due to two factors. First, the agent operates under severe partial observability: it receives no step-by-step instructions and has no access to an environment map, thus relying solely on a single egocentric RGB sensor to infer where to go. Second, real-world indoor scenes exhibit complex spatial structures, and long-horizon episodes often induce large discrepancies between the current observation and the goal image (e.g., in different rooms with little visual overlap). 
Together, these challenges leave the agent with little meaningful clues, 
causing it to drift into inefficient behaviors such as backtracking or aimless wandering.
Therefore, the key to improving navigation is to mine informative cues from the limited inputs to support reliable decision-making.

To realize that, prior works have largely followed two lines of research. 
\textbf{On the one hand}, modular-based methods~\cite{krantz2023iterative, lei2024instance} explicitly decompose the problem into several isolated tasks, and introduce additional sensors (e.g., depth and pose) to compensate for the lack of information. While effective, such pipelines typically incur extra hardware and system complexity, and their performance can be brittle due to error accumulation across modules. \textbf{On the other hand}, the end-to-end reinforcement learning~(RL) paradigm~\cite{sun2024fgprompt, li2025regnav} learns a navigation policy to directly map the visual representations
to the action control. Although RL has proven effective across multiple navigation tasks~\cite{huang2025mobilevla,qi2025vln,li2025regnav}, training an end-to-end policy purely with RL from scratch often fails to acquire strong spatial understanding and high-level planning capability, especially in unseen environments. 

Can we realize robust informative cue extraction without introducing additional sensors? One useful perspective comes from the fast-slow view of human cognition, which characterizes behavior as the interplay between a fast system~1 for reactive execution and a slow system~2 for deliberate reasoning~\cite{kahneman2011fast}. When the goal is distant and visual evidence is weak, humans often invoke the slow system to integrate sparse observations into higher-level spatial hypotheses and to sketch a coarse plan (e.g., exit the room), which in turn constrains subsequent action execution by the fast system. This division of labor effectively mines informative cues from limited inputs by converting ambiguous, partial observations into structured, short-horizon objectives, thereby reducing the search space and mitigating aimless wandering during long-horizon navigation.

Inspired by this observation, we aim to equip image-goal navigation agent with a similar high-level planning~(slow) and low-level execution~(fast) hierarchical mechanism. 
For the fast system, we train a lightweight navigation policy to directly output low-level actions for reactive control.
For the slow system, considering the strength in multi-modal understanding and reasoning, Vision-Language Models~(VLMs) are well-suited for this role. However, according to our experiments, simply zero-shotting the VLM yields limited gains. We attribute this to the fact that navigation videos primarily involve viewpoint changes and geometric transitions, which are less aligned with the event-centric understanding that VLMs are typically trained for. As a result, fully unlocking the potential of VLM-based slow planning requires task-specific finetuning. Yet, such finetuning relies on large-scale annotated short-horizon planning data, which is absent in standard image-goal navigation datasets. 
To bridge this gap, we collect a new  Hierarchical Reasoning dataset with annotated short-horizon goals for 767k trajectories.
This helps the VLM-based slow system to have substantially improved planning capability.

In this paper, we propose a Hierarchical Reasoning Navigation framework~(HRNav) to translate image-goal navigation to the interplay between high-level reasoning and low-level execution. Specifically, we adopt a two-stage training scheme: 1) Train the high-level reasoning with our self-collected planning dataset. 2) Freeze the slow system, incorporate its short-horizon planning ability to learn an efficient navigation policy with a novel Wandering Suppression Penalty~(WSP) to further reduce the wandering problem. 

We conclude our main contributions below:
\begin{itemize}
    \item We explore the aimless wandering problem in image-goal navigation and propose HRNav to combine VLM-based short-horizon reasoning with policy-level execution to map visual observations into actionable navigation intent.
    \item We collect a large planning dataset with 767k trajectories to train a VLM-based slow system with strong spatial understanding and reasoning ability. We also devise a novel penalty to further reduce the wandering problem.
    \item HRNav outperforms all the other state-of-the-art~(SOTA) methods in both simulated and real-world environments, offering a new pipeline in image-goal navigation and paving the way for future research in the field.
\end{itemize}

\begin{figure*}[t]
    \centering
    \includegraphics[width=\linewidth]{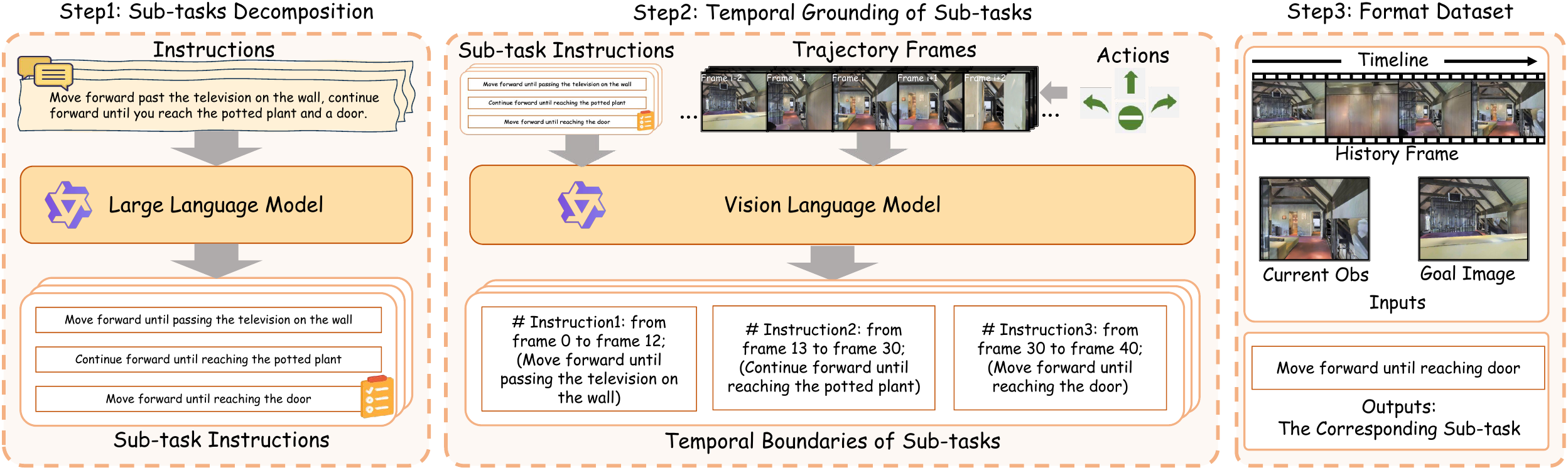}
    \caption{Overview of the hierarchical reasoning dataset construction pipeline.}
    \label{fig:dataset_construction}
    \vspace{-0.2cm}
\end{figure*}

\section{Related Work}
\subsection{Image-goal Navigation}
The image-goal navigation task requires the agent to perceive, reason, and execute actions in an unseen environment. Early modular-based methods~\cite{krantz2023iterative, lei2024instance} tackle this task step by step, which decomposes navigation into different perception, mapping, and path planning modules. To localize the agent, they usually utilize the mapping module to construct a map~\cite{yin2025unigoal,guo2025igl} or graph~\cite{kim2023topological, jiao2025litevloc} using additional sensors such as GPS, depth or pose information,  which limits their applicability in real-world robotic settings. An alternative line of work adopts end-to-end reinforcement learning paradigms, learning a navigation policy directly to map the goal image and observations to the low-level corresponding actions~\cite{al2022zero, sun2024fgprompt, li2025regnav}. While these methods demonstrate promising results, their training paradigms typically rely on training scenes and thus exhibit limited generalization to unseen scenes. Moreover, directly mapping the visual signal to low-level actions can be brittle in complex scenes, especially when the agent is far from the goal or located in a different room, often resulting in wandering behaviors. Recent work REGNav~\cite{li2025regnav} introduces room-level relationships into the navigation process to alleviate this issue. However, it lacks executable sub-task planning, and the navigation policy is still driven by visual matching rather than structured reasoning.

\subsection{Navigation with MLLM}
Recent advancements in multi-modal large language model~(MLLM)~\cite{zhang2024video,liu2024grounding,yang2025qwen3} have catalyzed progress across a wide range of research domains~\cite{guo2024llava,yuan2025videorefer,fu2024understanding,qi2026patchcue,liu2025mind,liu2024semantic,wu2025cemnetcrossemotionmemorynetwork,wu2025att}. Their strong multi-modal understanding and reasoning capabilities provide new opportunities for embodied navigation. One line of work leverages powerful closed-source models for zero-shot navigation within modular frameworks. These methods~\cite{yin2024sg,long2024instructnav,lyu2026himemvln} typically employ VLMs to perceive the environment and LLMs as planners to generate step-by-step decisions. For example, InstructNav~\cite{long2024instructnav} utilizes GPT-4V~\cite{yang2023dawn} to construct intuition value map and GPT-4~\cite{achiam2023gpt} for dynamic chain-of-navigation reasoning. SG-Nav~\cite{yin2024sg} and Unigoal~\cite{yin2025unigoal} employ LLaVA~\cite{liu2023visual} to construct an online 3D scene graph to prompt LLMs. While these methods achieve strong reasoning capability, their reliance on closed-source LLMs incurs high inference cost and limits practical deployment. Another line of work focuses on finetuning the open-source models on expert trajectories collected from simulator datasets. They usually adopt Video-based LLMs~\cite{lin2024vila,li2024llama,zhang2024video} to capture visual information and predict low-level action in an end-to-end manner. StreamVLN~\cite{wei2025streamvln} introduces an efficient framework that supports streaming visual inputs with bounded memory for action generation. CompassNav~\cite{li2025compassnav} further explores a two-stage training paradigm by combining supervised finetuning with reinforcement learning to improve navigation performance.
However, it remains challenging to finetune a VLM for image-goal navigation. Fully supervised learning will largely limit the exploration capability of the model, which is critical for the task. Instead of finetuning the VLM to predict actions, we teach it to make short-horizon plans. In this way, we leverage the VLM’s understanding and reasoning for high-level planning, while retaining the action policy’s capacity for efficient exploration.


\section{Method}
\begin{figure*}[t]
    \centering
    \includegraphics[width=\linewidth]{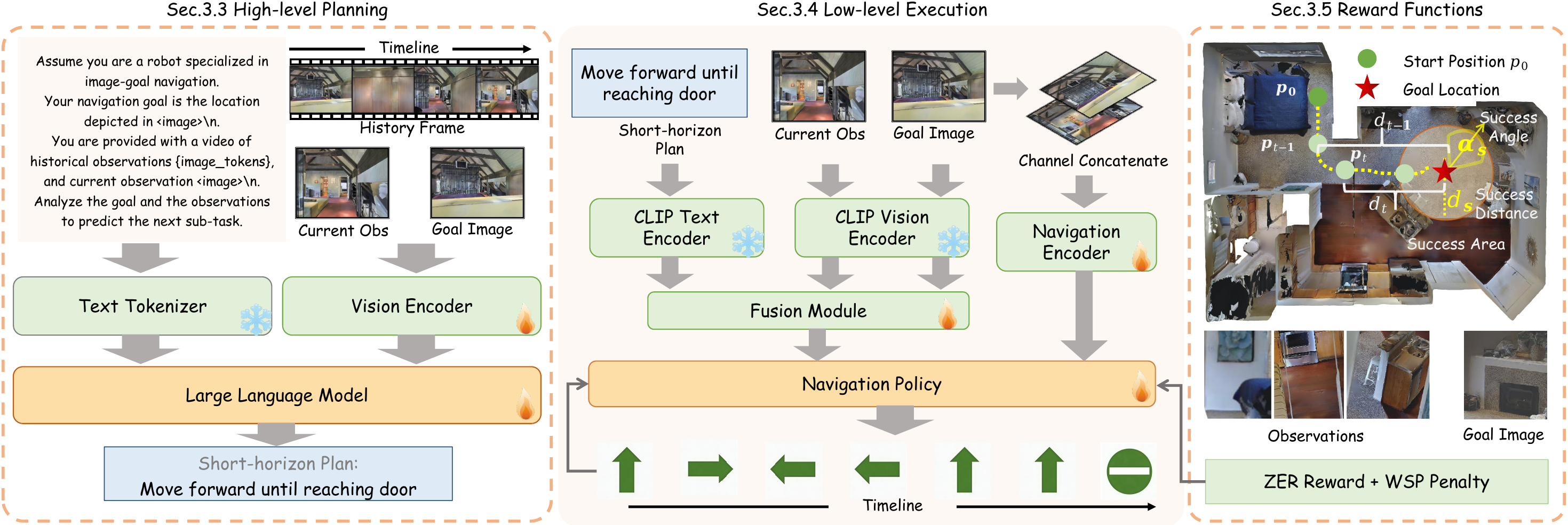}
    \caption{An overview of HRNav framework, where a high-level vision-language model predicts a short-horizon objective and a low-level policy executes actions via reinforcement learning under the reward functions.}
    \label{fig:framework}
    \vspace{-3mm}
\end{figure*}
\subsection{Task Setup}
In a navigation episode, the agent is initially placed at a random starting position $p_0$ in an unseen environment and is provided only with the goal image~$I_g$. At each time step $t$, the agent perceives the environment through an egocentric RGB observation~$I_o^t$ and selects a low-level action $a_t$ from the action space (e.g., move forward, turn left, turn right, and stop) and interacts with the environment, resulting in a new state and observation. The navigation episode terminates when the agent issues the stop action or reaches the maximum time steps. 

To tackle this task, we formulate image-goal navigation as a high-level planning and low-level execution problem. Specifically, a high-level planning module infers coarse-grained navigation intents based on the goal image and visual observations, while a low-level policy executes fine-grained actions conditioned on the inferred intents through reinforcement learning. 

\subsection{Hierarchical Reasoning Dataset}
\label{sec:Hierarchical Reasoning Dataset}
As illustrated in Fig.~\ref{fig:dataset_construction}, we construct a hierarchical reasoning dataset from existing VLN trajectory data (e.g., R2R-CE~\cite{krantz2020beyond} and RxR-CE~\cite{ku2020room}) to support sub-task planning based on visual observations. In these datasets, each trajectory consists of a natural language instruction that provides detailed instructions and a first-person video sequence that records the agent’s observation and execution in the environment.

\paragraph{Sub-task Decomposition.}
Given a full navigation instruction, we employ a large language model (Qwen3-14B~\cite{yang2025qwen3}) to decompose it into a temporally ordered sequence of non-overlapping, atomic sub-task instructions. 

\paragraph{Temporal Grounding of Sub-tasks.}
To align the decomposed sub-tasks with visual observations, we perform temporal grounding on the trajectory video.
Specifically, given the trajectory video and the ordered sub-task instructions, we prompt a vision-language model (Qwen2.5-VL-32B~\cite{bai2025qwen2}) to identify the temporal boundaries of each sub-task.
We augment each video frame with overlaid textual metadata, including the timestamp and the action executed at that time step.
The output is formatted as explicit frame intervals for each sub-task, which associate every video frame with its corresponding navigation sub-task.


\paragraph{Dataset Formatting.}
For each navigation trajectory, the last frame is selected as the goal image $I_g$. 
Given a current observation at time step $t$, we assign its supervision label as the sub-task that becomes active within a short future temporal window, encouraging the model to anticipate upcoming navigation intent. 
Each sample is formatted as
\emph{(history observations, current observation, goal image)} $\rightarrow$ \emph{next sub-task instruction}.
Running this pipeline produces 198K training samples from RxR~\cite{krantz2020beyond}, 239K from xsR2R~\cite{ku2020room}, and 330K from Youtube Videos~\cite{lin2023learning,cheng2024navila}, which serve as the training corpus for high-level planning. We further apply a Triple Quality Control Mechanism (TQCM) to filter malformed, temporally inconsistent, and semantically misaligned annotations. More details are provided in Appendix~\ref{app:dataset_quality}.

\begin{table*}[t]
\centering
\setlength{\tabcolsep}{4pt}
\renewcommand\arraystretch{1}
\begin{tabular}{c|cc|cc|cc|cc}
\hline
\multirow{2}{*}{\textbf{Method}}
&\multicolumn{2}{|c|}{\textbf{Easy}}
&\multicolumn{2}{|c|}{\textbf{Medium}}
&\multicolumn{2}{|c|}{\textbf{Hard}}
&\multicolumn{2}{|c}{\textbf{Overall}}\\
\cline{2-9}
& SR$\uparrow$ & SPL$\uparrow$
& SR$\uparrow$ & SPL$\uparrow$
& SR$\uparrow$ & SPL$\uparrow$
& SR$\uparrow$ & SPL$\uparrow$\\
\hline
VGM~\cite{kwon2021visual}&86.1\%&79.6\%&81.2\%&68.2\%&60.9\%&45.6\%&76.1\%&64.5\%\\
Mem-Aug~\cite{mezghan2022memory}&78.0\%&63.0\%&70.0\%&57.0\%&60.0\%&48.0\%&69.3\%&56.0\%\\
TSGM~\cite{kim2023topological}&91.1\%&\textbf{83.5\%}&82.0\%&68.1\%&70.3\%&50.0\%&81.1\%&67.2\%\\
FGPrompt-EF~\cite{sun2024fgprompt}&97.1\%&70.7\%&94.7\%&67.6\%&82.3\%&56.7\%&90.4\%&66.5\%\\
RFSG~\cite{feng2025image}&-&-&-&-&-&-&91.0\%&67.8\%\\
NavigateDiff~\cite{qin2025navigatediff}&-&-&-&-&-&-&91.0\%&64.8\%\\
REGNav~\cite{li2025regnav}&\underline{97.5\%}&71.4\%&\underline{95.4\%}&\underline{69.4\%}&\underline{87.1\%}&\underline{59.4\%}&\underline{92.9\%}&67.1\%\\

HRNav (This work)&\textbf{98.5\%}&\underline{75.2\%}&\textbf{96.2\%}&\textbf{73.7\%}&\textbf{87.2\%}&\textbf{66.8\%}&\textbf{94.0\%}&\textbf{71.2\%}\\
\hline
\end{tabular}
\caption{Comparison with state-of-the-art methods across three different difficulty levels on Gibson. The \textbf{best} and \underline{second} performance results are highlighted.}
\vspace{-2mm}
\label{tab:sota-gibson-memo}
\end{table*}

\subsection{High-level Planning}
\label{sec:hl}

The high-level planning module aims to infer the short-horizon plan based on visual observations and the goal image. We adopt VILA~\cite{lin2024vila} as the backbone, which consists of three main components: a vision encoder, a projector, and a large language model. 
Given a sequence of historical observations, the current observation, and the goal image, the vision encoder converts the input images into visual tokens, which are downsampled and mapped into the language domain through an MLP projector.
The projected visual tokens are concatenated with textual prompt tokens and fed into the LLM, which performs auto-regressive generation to predict the current short-horizon plan.

The high-level module is trained via supervised finetuning (SFT).
Besides our hierarchical reasoning dataset, we incorporate several auxiliary datasets.
Following NaVILA~\cite{cheng2024navila}, we include trajectory summarization data constructed from navigation videos~(EnvDrop~\cite{tan2019learning}), real-world 3D-grounded question answering data~(ScanQA~\cite{azuma2022scanqa}), and general VQA datasets~(ShareGPT-4V~\cite{chen2024sharegpt4v}, Video-chatgpt~\cite{maaz2024video}) to enhance the model’s scene understanding and multimodal reasoning ability.


\subsection{Low-level Execution}
\label{sec:ll}

Conditioned on the short-horizon objective generated by the high-level planning module, the low-level execution module produces executable navigation actions.
Inspired by PSL~\cite{sun2025prioritized}, we construct two complementary representations to capture both semantic intent and navigation-specific visual cues. First, the short-horizon plan, current observation, and goal image are encoded using CLIP~\cite{radford2021learning} text and vision encoders, respectively.
The resulting features are fused through a multimodal fusion module to obtain a semantic representation.
Second, to preserve spatial information for navigation, the current observation and the goal image are concatenated along the channel dimension and processed by a navigation encoder that focuses on geometry and layout cues. Then, we concatenate the semantic and navigation features as the fused features.

The fused features $f_{fused}$ and previous actions~$a_{t-1}$ are jointly fed into a navigation policy network~$\pi$ to predict the agent's current state embedding $s_t$ at time $t$, which is defined as:
\begin{equation}
s_t = \pi(f_{fused} \oplus a_{t-1} \mid h_{t-1}),
\end{equation}
where $h_{t-1}$ is the hidden layer of the recurrent layers in the policy from the previous step. An actor-critic network then utilizes $s_t$ to predict the state value and determine the agent’s next action. 
The low-level policy is optimized using reinforcement learning, allowing it to maintain exploration capability and adapt to unseen environments. 
More policy details can be found in Appendix Sec~\ref{navigation policy}.

\subsection{Reward Functions}
\label{sec:reward}

\paragraph{ZER Reward.}
Following ZER~\cite{al2022zero}, we use a dense distance-and-view shaping reward with a sparse success reward:
\vspace{-2mm}
\begin{equation}
r_t = (d_{t-1}-d_t) + \mathbb{I}(d_t\le d_s)(\alpha_{t-1}-\alpha_t) - \gamma,
\label{eq:reward_zer}
\end{equation}
\begin{equation}
R_s = 5\Big[\mathbb{I}(d_t\le d_s) + \mathbb{I}(d_t\le d_s \wedge \alpha_t\le \alpha_s)\Big],
\label{eq:success_zer}
\end{equation}
\vspace{-1mm}
where $d_t$ is the geodesic distance, $\alpha_t$ is the view-angle difference, and the view shaping is activated only within the success area ($d_t\le d_s$).
We set $d_s{=}1$m and $\alpha_s{=}25^\circ$.
However, the shaping term is purely local and does not explicitly penalize redundant motions, making the agent prone to short-range oscillations and backtracking.

\paragraph{Wandering Suppression Penalty.}
Built upon the above reward, we further introduce a \emph{Wandering Suppression Penalty} (WSP) that explicitly discourages unnecessary detours and short-term revisits.
Concretely, WSP consists of a path-length penalty and a revisit penalty:
\vspace{-3mm}
\begin{equation}
r_t^{\text{wsp}}
=
-(\ell_t-\ell_{t-1})
-\Delta c_t,
\label{eq:wsp}
\vspace{-3mm}
\end{equation}
where $\ell_t$ is the cumulative traveled path length, $(\ell_t-\ell_{t-1})$ is the step displacement, $c_t$ is a monotonic oscillation counter that increases when the agent exhibits short-term backtracking. The overall step reward is:
\vspace{-3mm}
\begin{equation}
\tilde{r}_t = r_t + R_s + \lambda_w  r_t^{\text{wsp}},
\label{eq:reward_full_wsp}
\vspace{-3mm}
\end{equation}
where $\lambda_w$ is the weight hyperparameter. The ablation on $\lambda_w$ can be found in Appendix Sec.~\ref{ablation on weights}. More reward details can be found in Appendix Sec.~\ref{zer reward} and Sec.~\ref{wsp}. 

\section{Experiments}
\subsection{Simulation Experiments}
\paragraph{Datasets and Evaluation Metrics.}
 The high-level planning module is trained via supervised finetuning using the datasets described in Sec.~\ref{sec:hl}, including our hierarchical reasoning dataset as well as auxiliary navigation and visual reasoning datasets. For the low-level execution module, all of the experiments are conducted on the Habitat simulator~\cite{savva2019habitat,szot2021habitat}. We train it on the Gibson dataset~\cite{xiazamirhe2018gibsonenv} using the dataset split provided by~\cite{mezghan2022memory}. Gibson contains diverse indoor environments and consists of 9000 training episodes from 72 scenes and 4200 testing episodes from 14 scenes. To evaluate cross-domain generalization, we test the trained agent on the Matterport3D (MP3D)~\cite{chang2017matterport3d} and Habitat-Matterport3D (HM3D)~\cite{ramakrishnan2021habitat} datasets.

\begin{table}[t]
\centering
\setlength{\tabcolsep}{6pt}
\renewcommand\arraystretch{1}
\begin{tabular}{c|cc}
\hline
\multirow{2}{*}{\textbf{Method}}&\multicolumn{2}{|c}{\textbf{MP3D}}\\
\cline{2-3}&SR$\uparrow$&SPL$\uparrow$\\
\hline
Mem-Aug&6.9\%&3.9\%\\
ZER&14.6\%&10.8\%\\
FGPrompt-EF&75.7\%&48.8\%\\
REGNav&\underline{78.0\%}&\underline{50.2\%}\\
HRNav&\textbf{81.4\%}\small(+3.4\%)&\textbf{56.3\%}\small(+6.1\%)\\
\hline
\end{tabular}
\caption{Cross-domain evaluation on MP3D.}
\label{tab:cross-eval-mp3d}
\end{table}

We adopt standard navigation metrics including Success Rate (SR) and Success weighted by Path Length (SPL)~\cite{anderson2018evaluation}. SPL balances the efficiency and success rate by calculating the weighted sum of the ratio of the shortest path length to the predicted path length. The maximum number of steps per episode is set to 500.


\begin{figure*}[t]
    \centering
    \includegraphics[width=\linewidth]{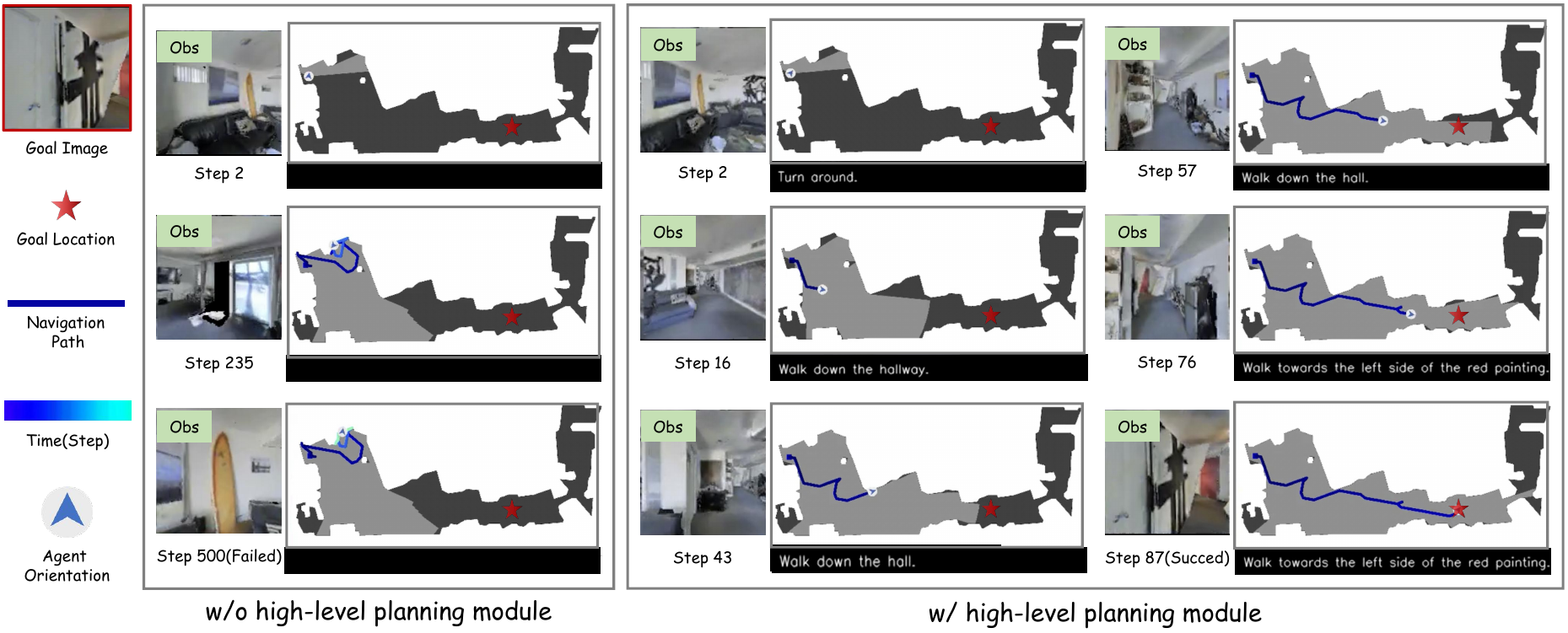}
    \caption{Visualization of navigation trajectories with and without high-level planning. For the model with the high-level planning module (right), each top-view map is annotated with the short-horizon objective, which is predicted online and remains consistent with the subsequent navigation path.
    High-level planning leads to more efficient and successful navigation.}
    \label{fig:vis_path}
\end{figure*}

\paragraph{Implementation Details}
We adopt a two-stage training scheme: first train the high-level module with supervised finetuning (SFT) for one epoch on the curated corpus.
After SFT, the high-level module is frozen and only used for inference; then train the low-level policy with DD-PPO~\cite{wijmans2019dd}.
We keep the same \emph{fast--slow} schedule: the low-level policy runs frequently to output actions, while the high-level module is invoked sparsely to update sub-task instructions (every 15 steps). 
In simulation, the agent height is 1.5m with a radius of 0.1m, using a single RGB sensor with a $90^{\circ}$ FOV at $128{\times}128$ resolution.
The action space includes MOVE\_FORWARD (0.25m), TURN\_LEFT/RIGHT ($30^{\circ}$), and STOP.
We set the slack reward $\gamma$ to $0.01$.
All experiments are conducted on 8 NVIDIA H20 GPUs; high-level finetuning takes 64 hours and low-level training takes 30 hours~(20M steps).

\paragraph{Efficiency Analysis.} HRNav achieves 41.16\,ms average latency per step and 24.29 FPS over 5,000 navigation steps on one NVIDIA H20 GPU with the default 15-step planning interval. Although each slow-planner call takes 374.12\,ms, its cost is amortized by sparse invocation, while the low-level executor runs at 14.22\,ms per step. Detailed results are given in Appendix~\ref{app:efficiency}.

\paragraph{In-domain Evaluation on Gibson.} Table~\ref{tab:sota-gibson-memo} reports the performance comparison on the Gibson dataset. (We report the results averaged over 3 random seeds. The variances are less than 1e-3.)
HRNav achieves the best overall results, reaching an SR of 94.0\% and an SPL of 71.2\%.
Compared with all prior methods, HRNav improves the overall SPL by \textbf{+4.1\%} and the SR by \textbf{+1.1\%}.
Notably, the gains are more pronounced on challenging scenarios: HRNav improves SPL by \textbf{+4.3\%} on the Medium set and \textbf{+7.4\%} on the Hard set, demonstrating more efficient long-horizon navigation and reduced wandering behavior. 

\begin{table}[t]
\centering
\setlength{\tabcolsep}{6pt}
\renewcommand\arraystretch{1}
\begin{tabular}{c|cc}
\hline
\multirow{2}{*}{\textbf{Method}}&\multicolumn{2}{|c}{\textbf{HM3D}}\\
\cline{2-3}&SR$\uparrow$&SPL$\uparrow$\\
\hline
Mem-Aug&3.5\%&1.9\%\\
ZER&9.6\%&6.3\%\\
FGPrompt-EF&\underline{75.2\%}&42.1\%\\
RFSG&73.4\%&42.7\%\\
REGNav&\underline{75.2\%}&\underline{44.0\%}\\
HRNav &\textbf{80.0\%}\small(+4.8\%)&\textbf{49.3\%}\small(+5.4\%)\\
\hline
\end{tabular}
\caption{Cross-domain evaluation on HM3D.}
\label{tab:cross-eval-hm3d}
\end{table}

\paragraph{Cross Domain Evaluation on MP3D and HM3D.}
Tables~\ref{tab:cross-eval-mp3d} and~\ref{tab:cross-eval-hm3d} report cross-domain evaluation results on MP3D and HM3D.
All methods are trained on the Gibson dataset and directly tested on these datasets without any finetuning, evaluating their generalization ability to unseen environments.

HRNav achieves the best performance on both benchmarks.
On MP3D, HRNav reaches an SR of 81.4\% and an SPL of 56.3\%, outperforming the strongest prior method REGNav by \textbf{+3.4\% SR} and \textbf{+6.1\% SPL}.
On the more challenging HM3D benchmark, HRNav further improves SR to 80.0\% and SPL to 49.3\%, exceeding REGNav by \textbf{+4.8\% SR} and \textbf{+5.4\% SPL}.
These consistent gains demonstrate that hierarchical short-horizon planning significantly improves cross-domain robustness, enabling more efficient navigation trajectories and reducing domain-specific overfitting.

\paragraph{Visualization.}
Fig.~\ref{fig:vis_path} presents a representative visualization comparing navigation behaviors with and without the high-level planning module.
Without high-level planning, the agent exhibits frequent backtracking and wandering, eventually failing to reach the goal within the maximum step budget.
In contrast, with short-horizon objectives, the agent follows a more direct and coherent trajectory, successfully reaching the goal with fewer redundant explorations.
This example qualitatively demonstrates how high-level planning provides effective guidance for long-horizon navigation and improves trajectory efficiency.
Additional qualitative results are provided in the appendix.

\begin{table}[t]
\centering
\setlength{\tabcolsep}{9pt}
\renewcommand\arraystretch{1}
\begin{tabular}{l|cc}
\hline
\textbf{Method} & \textbf{SR$\uparrow$} & \textbf{SPL$\uparrow$} \\
\hline
HRNav &\textbf{93.36\%}&\textbf{66.21\%}\\
\hline
\multicolumn{3}{l}{\textit{Ablation A: High-level planning}} \\
w/o HL & 92.52\% & 63.53\% \\
rp HL with Qwen2& 81.62\%& 55.29\% \\
\hline
\multicolumn{3}{l}{\textit{Ablation B: Low-level policy}} \\
Semantic-only& 49.4\% & 32.6\% \\
Nav-only& 81.6\% & 61.1\% \\
w/o prev actions& 87.7\%&  62.3\%  \\
\hline
\end{tabular}
\caption{Ablation study on different components on Gibson test set. All the experiments are conducted without the new reward. }
\vspace{-3mm}
\label{tab:ablation_components_gibson}
\end{table}

\begin{figure*}[t]
    \centering
    \includegraphics[width=\linewidth]{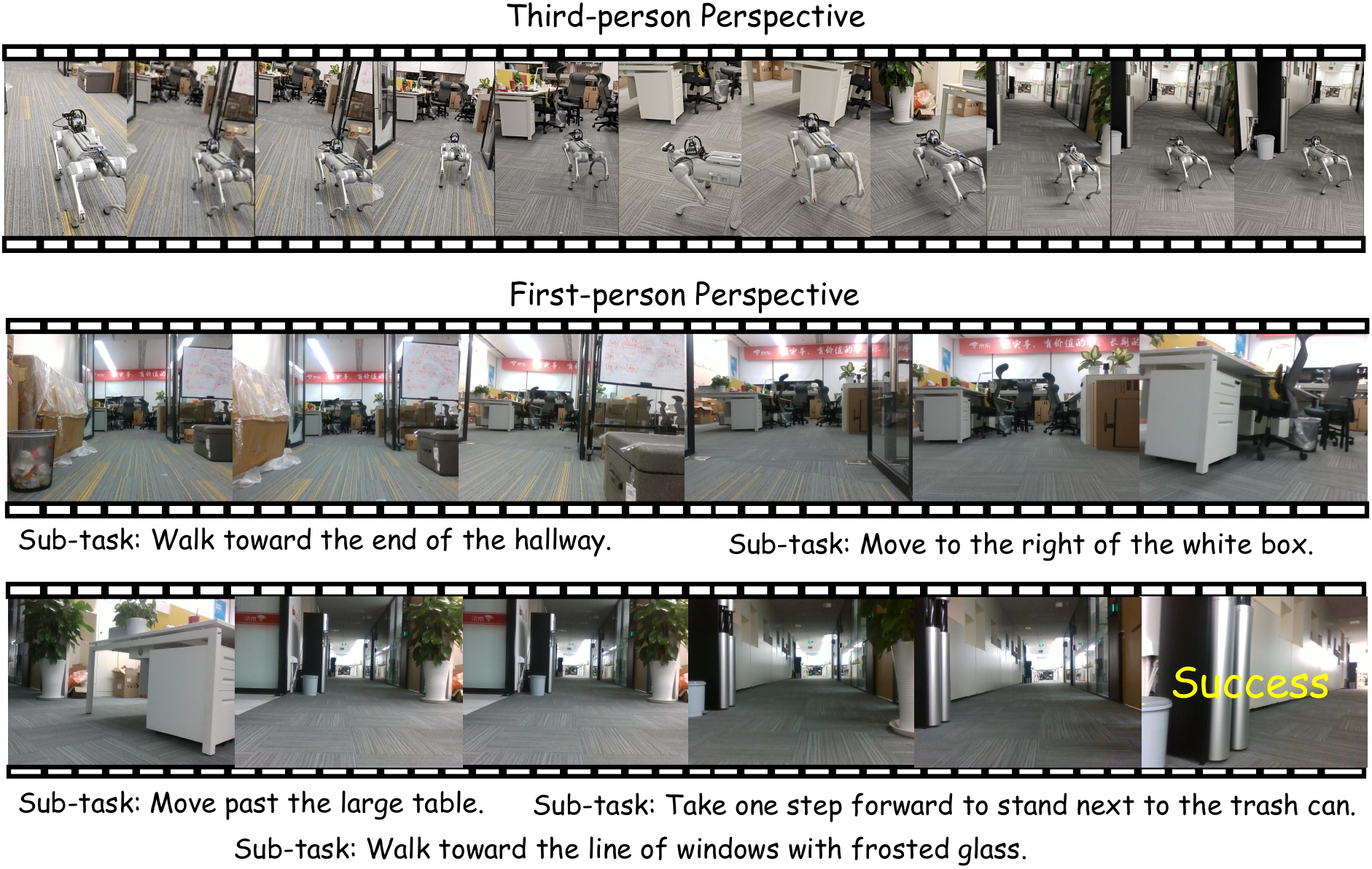}
    \caption{Real-world qualitative visualization of HRNav.}
    \label{fig:real_world_vis}
\end{figure*}

\paragraph{Ablation Studies.}
Table~\ref{tab:ablation_components_gibson} summarizes the ablation results on model components.
Removing the high-level planning module (w/o HL) leads to a clear performance drop in both SR and SPL, indicating that a hierarchical short-horizon objective is essential for effective navigation.
Replacing the high-level module with Qwen2 also degrades performance compared to the full model, suggesting that the proposed high-level design provides more suitable objectives for this task.
For the low-level policy, removing either semantic features (Semantic-only) or navigation-specific visual features (Nav-only) causes substantial degradation, with the semantic-only variant failing dramatically.
This demonstrates that both semantic intent and structural navigation cues are necessary.
In addition, excluding previous actions reduces performance, highlighting the importance of action history for stable control.

\begin{table}[t]
\centering
\setlength{\tabcolsep}{10pt}
\renewcommand\arraystretch{1}
\begin{tabular}{l|cc}
\hline
\textbf{Setting} & \textbf{SR$\uparrow$} & \textbf{SPL$\uparrow$} \\
\hline
w/o new reward & 93.36\% & 66.21\% \\
only revisit & 91.62\% & 70.62\% \\
only path\_len & 91.43\% & 68.13\% \\
revisit+path\_len & \textbf{94.00\%} & \textbf{71.15\%} \\
\hline
\end{tabular}
\caption{Ablation study on the proposed reward design on Gibson test set.}
\vspace{-3mm}
\label{tab:ablation_reward_gibson}
\end{table}

Table~\ref{tab:ablation_reward_gibson} evaluates the impact of the proposed reward components.
Without the new reward, the agent achieves an SPL of 66.21\%.
Introducing either the revisit penalty or the path-length regularization alone improves SPL, while combining both yields the best performance, improving SPL to \textbf{71.15\%} and SR to \textbf{94.00\%}.
These results show that the two reward components are complementary and jointly encourage more efficient and less redundant navigation trajectories.

\paragraph{Effectiveness of High-Level Planning Module.} To more intuitively assess whether our slow system can perform high-level planning accurately, we randomly sampled 100 trajectories from our collected dataset for evaluation. We feed the history and observation images into different models in the same manner and use the same prompt to elicit short-horizon planning. We compare with Qwen2-VL~\cite{bai2025qwen2} and Qwen3-VL~\cite{yang2025qwen3}, which are two widely used open-source VLMs. Four metrics~\cite{wu2025event} are calculated: BLEU-4~\cite{papineni2002bleu}, METEOR~\cite{banerjee2005meteor}, ROUGE-L~\cite{lin2004rouge}, and CIDEr~\cite{vedantam2015cider}. The results are presented in TABLE~\ref{tab:eval_slow_system}. It's clear that after training, our model can generate short-horizon goals more precisely. This also helps explain why HRNav ultimately achieves stronger navigation performance.

\begin{table}[t]
\centering
\setlength{\tabcolsep}{9pt}
\renewcommand\arraystretch{1}
\begin{tabular}{l|cccc}
\hline
\textbf{Model} & \textbf{B4$\uparrow$} & \textbf{M$\uparrow$}& \textbf{C$\uparrow$}& \textbf{R$\uparrow$} \\
\hline
Qwen2-VL & 1.2 &9.1&0.13&22.8 \\
Qwen3-VL &3.9&14.4&0.23&21.8 \\
HRNav-Slow & \textbf{31.5} & \textbf{38.3}&\textbf{2.74}&\textbf{45.6} \\
\hline
\end{tabular}
\caption{Evaluation of high-level planning on a test subset of the self-collected Hierarchical Reasoning dataset.}
\vspace{-3mm}
\label{tab:eval_slow_system}
\end{table}

\subsection{Real-World Experiments}
\paragraph{Settings.} Our real-world experiments were conducted on a Unitree Go2 robot equipped with an Intel RealSense D435i camera on the front.
During navigation, the system continuously streams RGB observations from the D435i to a remote server powered by NVIDIA H20 GPUs.
Given a goal image, HRNav processes the observations to generate the next sub-task and predicts actions.
The predicted action is then transmitted back to the Go2, which executes it by calling the robot motion API.

\paragraph{Qualitative results.}
Fig.~\ref{fig:real_world_vis} visualizes a successful real-world navigation episode executed by our HRNav. The goal image depicts a water dispenser near the trash can. The robot is initially placed in the room adjacent to the goal.
As shown in the figure, the slow system guides the robot to exit the room by walking down the hallway, and then sets different short-horizon goals to move the robot closer to the target location. When the object in the goal image comes into view, the slow system accurately recognizes it and directs the robot to approach the trash can and stop. This demonstrates that by building a hierarchical fast–slow system, HRNav can reduce a difficult task into manageable steps and reach the destination progressively.

\section{Conclusion}
We propose \textbf{HRNav}, a hierarchical framework for image-goal navigation. 
Inspired by human cognition, we combine a slow planning module with a fast execution policy for efficient control.
To train the slow system, a hierarchical reasoning dataset is constructed by decomposing task instructions into temporally grounded sub-tasks.
We add a Wandering Suppression Penalty, which greatly reduces the wandering problem.
Experiments on Gibson and cross-domain MP3D/HM3D datasets demonstrate promising results, while ablations validate the effectiveness of key components.

\section*{Limitations}
Although HRNav demonstrates promising perfomance, sim-to-real transfer remains challenging. 
First, there is a noticeable gap between the camera settings in simulation and those on the physical robots, leading to mismatched viewpoints and visual distributions.
This discrepancy can distort the perceived appearance of rooms, landmarks, and obstacles, thereby degrading sub-task reasoning and action selection in real-world deployment.
Second, the robot’s body embodiment is not fully modeled during simulation training.
In practice, a quadrupedal robot may get stuck or experience unintended contacts when moving close to obstacles (e.g., its hind legs may scrape against furniture), which are not captured by the simulator.

In future work, we will try to incorporate more accurate sensor and embodiment modeling (e.g., camera configuration and full-body collision geometry) to narrow the sim-to-real gap and improve robustness on real robots.


\section*{Acknowledgement}
This work was supported by Fundamental and Interdisciplinary Disciplines Breakthrough Plan of the Ministry of Education of China under Grant JYB2025XDXM504.
\bibliography{custom}

@inproceedings{zheng2024towards,
  title={Towards learning a generalist model for embodied navigation},
  author={Zheng, Duo and Huang, Shijia and Zhao, Lin and Zhong, Yiwu and Wang, Liwei},
  booktitle={Proceedings of the IEEE/CVF Conference on Computer Vision and Pattern Recognition},
  pages={13624--13634},
  year={2024}
}

@inproceedings{gan2020look,
  title={Look, listen, and act: Towards audio-visual embodied navigation},
  author={Gan, Chuang and Zhang, Yiwei and Wu, Jiajun and Gong, Boqing and Tenenbaum, Joshua B},
  booktitle={2020 IEEE International Conference on Robotics and Automation (ICRA)},
  pages={9701--9707},
  year={2020},
  organization={IEEE}
}

@article{huang2025mobilevla,
  title={MobileVLA-R1: Reinforcing Vision-Language-Action for Mobile Robots},
  author={Huang, Ting and Li, Dongjian and Yang, Rui and Zhang, Zeyu and Yang, Zida and Tang, Hao},
  journal={arXiv preprint arXiv:2511.17889},
  year={2025}
}

@article{qi2025vln,
  title={VLN-R1: Vision-Language Navigation via Reinforcement Fine-Tuning},
  author={Qi, Zhangyang and Zhang, Zhixiong and Yu, Yizhou and Wang, Jiaqi and Zhao, Hengshuang},
  journal={arXiv preprint arXiv:2506.17221},
  year={2025}
}

@inproceedings{krantz2023navigating,
  title={Navigating to objects specified by images},
  author={Krantz, Jacob and Gervet, Theophile and Yadav, Karmesh and Wang, Austin and Paxton, Chris and Mottaghi, Roozbeh and Batra, Dhruv and Malik, Jitendra and Lee, Stefan and Chaplot, Devendra Singh},
  booktitle={Proceedings of the IEEE/CVF International Conference on Computer Vision},
  pages={10916--10925},
  year={2023}
}

@article{sun2024fgprompt,
  title={FGPrompt: fine-grained goal prompting for image-goal navigation},
  author={Sun, Xinyu and Chen, Peihao and Fan, Jugang and Chen, Jian and Li, Thomas and Tan, Mingkui},
  journal={Advances in Neural Information Processing Systems},
  volume={36},
  year={2024}
}

@article{majumdar2022zson,
  title={Zson: Zero-shot object-goal navigation using multimodal goal embeddings},
  author={Majumdar, Arjun and Aggarwal, Gunjan and Devnani, Bhavika and Hoffman, Judy and Batra, Dhruv},
  journal={Advances in Neural Information Processing Systems},
  volume={35},
  pages={32340--32352},
  year={2022}
}

@inproceedings{al2022zero,
  title={Zero experience required: Plug \& play modular transfer learning for semantic visual navigation},
  author={Al-Halah, Ziad and Ramakrishnan, Santhosh Kumar and Grauman, Kristen},
  booktitle={Proceedings of the IEEE/CVF Conference on Computer Vision and Pattern Recognition},
  pages={17031--17041},
  year={2022}
}

@inproceedings{mezghan2022memory,
  title={Memory-augmented reinforcement learning for image-goal navigation},
  author={Mezghan, Lina and Sukhbaatar, Sainbayar and Lavril, Thibaut and Maksymets, Oleksandr and Batra, Dhruv and Bojanowski, Piotr and Alahari, Karteek},
  booktitle={2022 IEEE/RSJ International Conference on Intelligent Robots and Systems (IROS)},
  pages={3316--3323},
  year={2022},
  organization={IEEE}
}

@inproceedings{kim2023topological,
  title={Topological semantic graph memory for image-goal navigation},
  author={Kim, Nuri and Kwon, Obin and Yoo, Hwiyeon and Choi, Yunho and Park, Jeongho and Oh, Songhwai},
  booktitle={Conference on Robot Learning},
  pages={393--402},
  year={2023},
  organization={PMLR}
}

@inproceedings{zhu2017target,
  title={Target-driven visual navigation in indoor scenes using deep reinforcement learning},
  author={Zhu, Yuke and Mottaghi, Roozbeh and Kolve, Eric and Lim, Joseph J and Gupta, Abhinav and Fei-Fei, Li and Farhadi, Ali},
  booktitle={2017 IEEE international conference on robotics and automation (ICRA)},
  pages={3357--3364},
  year={2017},
  organization={IEEE}
}

@inproceedings{xiazamirhe2018gibsonenv,
  title={Gibson env: real-world perception for embodied agents},
  author={Xia, Fei and R. Zamir, Amir and He, Zhi-Yang and Sax, Alexander and Malik, Jitendra and Savarese, Silvio},
  booktitle={Computer Vision and Pattern Recognition (CVPR), 2018 IEEE Conference on},
  year={2018},
  organization={IEEE}
}

@inproceedings{wasserman2023last,
  title={Last-mile embodied visual navigation},
  author={Wasserman, Justin and Yadav, Karmesh and Chowdhary, Girish and Gupta, Abhinav and Jain, Unnat},
  booktitle={Conference on Robot Learning},
  pages={666--678},
  year={2023},
  organization={PMLR}
}

@inproceedings{lei2024instance,
  title={Instance-aware Exploration-Verification-Exploitation for Instance ImageGoal Navigation},
  author={Lei, Xiaohan and Wang, Min and Zhou, Wengang and Li, Li and Li, Houqiang},
  booktitle={Proceedings of the IEEE/CVF Conference on Computer Vision and Pattern Recognition},
  pages={16329--16339},
  year={2024}
}

@inproceedings{savva2019habitat,
  title={Habitat: A platform for embodied ai research},
  author={Savva, Manolis and Kadian, Abhishek and Maksymets, Oleksandr and Zhao, Yili and Wijmans, Erik and Jain, Bhavana and Straub, Julian and Liu, Jia and Koltun, Vladlen and Malik, Jitendra and others},
  booktitle={Proceedings of the IEEE/CVF international conference on computer vision},
  pages={9339--9347},
  year={2019}
}

@article{szot2021habitat,
  title={Habitat 2.0: Training home assistants to rearrange their habitat},
  author={Szot, Andrew and Clegg, Alexander and Undersander, Eric and Wijmans, Erik and Zhao, Yili and Turner, John and Maestre, Noah and Mukadam, Mustafa and Chaplot, Devendra Singh and Maksymets, Oleksandr and others},
  journal={Advances in neural information processing systems},
  volume={34},
  pages={251--266},
  year={2021}
}

@article{chang2017matterport3d,
  title={Matterport3d: Learning from rgb-d data in indoor environments},
  author={Chang, Angel and Dai, Angela and Funkhouser, Thomas and Halber, Maciej and Niessner, Matthias and Savva, Manolis and Song, Shuran and Zeng, Andy and Zhang, Yinda},
  journal={arXiv preprint arXiv:1709.06158},
  year={2017}
}

@article{ramakrishnan2021habitat,
  title={Habitat-matterport 3d dataset (hm3d): 1000 large-scale 3d environments for embodied ai},
  author={Ramakrishnan, Santhosh K and Gokaslan, Aaron and Wijmans, Erik and Maksymets, Oleksandr and Clegg, Alex and Turner, John and Undersander, Eric and Galuba, Wojciech and Westbury, Andrew and Chang, Angel X and others},
  journal={arXiv preprint arXiv:2109.08238},
  year={2021}
}

@article{anderson2018evaluation,
  title={On evaluation of embodied navigation agents},
  author={Anderson, Peter and Chang, Angel and Chaplot, Devendra Singh and Dosovitskiy, Alexey and Gupta, Saurabh and Koltun, Vladlen and Kosecka, Jana and Malik, Jitendra and Mottaghi, Roozbeh and Savva, Manolis and others},
  journal={arXiv preprint arXiv:1807.06757},
  year={2018}
}

@inproceedings{kwon2021visual,
  title={Visual graph memory with unsupervised representation for visual navigation},
  author={Kwon, Obin and Kim, Nuri and Choi, Yunho and Yoo, Hwiyeon and Park, Jeongho and Oh, Songhwai},
  booktitle={Proceedings of the IEEE/CVF international conference on computer vision},
  pages={15890--15899},
  year={2021}
}

@inproceedings{he2016deep,
  title={Deep residual learning for image recognition},
  author={He, Kaiming and Zhang, Xiangyu and Ren, Shaoqing and Sun, Jian},
  booktitle={Proceedings of the IEEE conference on computer vision and pattern recognition},
  pages={770--778},
  year={2016}
}

@inproceedings{deng2009imagenet,
  title={Imagenet: A large-scale hierarchical image database},
  author={Deng, Jia and Dong, Wei and Socher, Richard and Li, Li-Jia and Li, Kai and Fei-Fei, Li},
  booktitle={2009 IEEE conference on computer vision and pattern recognition},
  pages={248--255},
  year={2009},
  organization={Ieee}
}

@inproceedings{krantz2023iterative,
  title={Iterative vision-and-language navigation},
  author={Krantz, Jacob and Banerjee, Shurjo and Zhu, Wang and Corso, Jason and Anderson, Peter and Lee, Stefan and Thomason, Jesse},
  booktitle={Proceedings of the IEEE/CVF Conference on Computer Vision and Pattern Recognition},
  pages={14921--14930},
  year={2023}
}

@article{wijmans2019dd,
  title={Dd-ppo: Learning near-perfect pointgoal navigators from 2.5 billion frames},
  author={Wijmans, Erik and Kadian, Abhishek and Morcos, Ari and Lee, Stefan and Essa, Irfan and Parikh, Devi and Savva, Manolis and Batra, Dhruv},
  journal={arXiv preprint arXiv:1911.00357},
  year={2019}
}

@article{kingma2014adam,
  title={Adam: A method for stochastic optimization},
  author={Kingma, Diederik P},
  journal={arXiv preprint arXiv:1412.6980},
  year={2014}
}

@inproceedings{sun2025prioritized,
  title={Prioritized semantic learning for zero-shot instance navigation},
  author={Sun, Xinyu and Liu, Lizhao and Zhi, Hongyan and Qiu, Ronghe and Liang, Junwei},
  booktitle={European Conference on Computer Vision},
  pages={161--178},
  year={2025},
  organization={Springer}
}

@inproceedings{li2025regnav,
  title={REGNav: Room Expert Guided Image-Goal Navigation},
  author={Li, Pengna and Wu, Kangyi and Fu, Jingwen and Zhou, Sanping},
  booktitle={Proceedings of the AAAI Conference on Artificial Intelligence},
  volume={39},
  number={5},
  pages={4860--4868},
  year={2025}
}

@inproceedings{jiao2025litevloc,
  title={LiteVLoc: Map-lite visual localization for image goal navigation},
  author={Jiao, Jianhao and He, Jinhao and Liu, Changkun and Aegidius, Sebastian and Hu, Xiangcheng and Braud, Tristan and Kanoulas, Dimitrios},
  booktitle={2025 IEEE International Conference on Robotics and Automation (ICRA)},
  pages={5244--5251},
  year={2025},
  organization={IEEE}
}

@inproceedings{guo2025igl,
  title={IGL-Nav: Incremental 3D Gaussian Localization for Image-goal Navigation},
  author={Guo, Wenxuan and Xu, Xiuwei and Yin, Hang and Wang, Ziwei and Feng, Jianjiang and Zhou, Jie and Lu, Jiwen},
  booktitle={Proceedings of the IEEE/CVF International Conference on Computer Vision},
  pages={6808--6817},
  year={2025}
}

@inproceedings{yuan2025videorefer,
  title={Videorefer suite: Advancing spatial-temporal object understanding with video llm},
  author={Yuan, Yuqian and Zhang, Hang and Li, Wentong and Cheng, Zesen and Zhang, Boqiang and Li, Long and Li, Xin and Zhao, Deli and Zhang, Wenqiao and Zhuang, Yueting and others},
  booktitle={Proceedings of the Computer Vision and Pattern Recognition Conference},
  pages={18970--18980},
  year={2025}
}

@inproceedings{guo2024llava,
  title={LLaVA-ultra: Large Chinese language and vision assistant for ultrasound},
  author={Guo, Xuechen and Chai, Wenhao and Li, Shi-Yan and Wang, Gaoang},
  booktitle={Proceedings of the 32nd ACM international conference on multimedia},
  pages={8845--8854},
  year={2024}
}

@article{zhang2024video,
  title={Video instruction tuning with synthetic data},
  author={Zhang, Yuanhan and Wu, Jinming and Li, Wei and Li, Bo and Ma, Zejun and Liu, Ziwei and Li, Chunyuan},
  journal={arXiv preprint arXiv:2410.02713},
  year={2024}
}

@inproceedings{liu2024grounding,
  title={Grounding dino: Marrying dino with grounded pre-training for open-set object detection},
  author={Liu, Shilong and Zeng, Zhaoyang and Ren, Tianhe and Li, Feng and Zhang, Hao and Yang, Jie and Jiang, Qing and Li, Chunyuan and Yang, Jianwei and Su, Hang and others},
  booktitle={European conference on computer vision},
  pages={38--55},
  year={2024},
  organization={Springer}
}

@article{long2024instructnav,
  title={Instructnav: Zero-shot system for generic instruction navigation in unexplored environment},
  author={Long, Yuxing and Cai, Wenzhe and Wang, Hongcheng and Zhan, Guanqi and Dong, Hao},
  journal={arXiv preprint arXiv:2406.04882},
  year={2024}
}

@article{achiam2023gpt,
  title={Gpt-4 technical report},
  author={Achiam, Josh and Adler, Steven and Agarwal, Sandhini and Ahmad, Lama and Akkaya, Ilge and Aleman, Florencia Leoni and Almeida, Diogo and Altenschmidt, Janko and Altman, Sam and Anadkat, Shyamal and others},
  journal={arXiv preprint arXiv:2303.08774},
  year={2023}
}

@article{yang2023dawn,
  title={The dawn of lmms: Preliminary explorations with gpt-4v (ision)},
  author={Yang, Zhengyuan and Li, Linjie and Lin, Kevin and Wang, Jianfeng and Lin, Chung-Ching and Liu, Zicheng and Wang, Lijuan},
  journal={arXiv preprint arXiv:2309.17421},
  year={2023}
}

@article{yin2024sg,
  title={Sg-nav: Online 3d scene graph prompting for llm-based zero-shot object navigation},
  author={Yin, Hang and Xu, Xiuwei and Wu, Zhenyu and Zhou, Jie and Lu, Jiwen},
  journal={Advances in neural information processing systems},
  volume={37},
  pages={5285--5307},
  year={2024}
}

@inproceedings{yin2025unigoal,
  title={Unigoal: Towards universal zero-shot goal-oriented navigation},
  author={Yin, Hang and Xu, Xiuwei and Zhao, Linqing and Wang, Ziwei and Zhou, Jie and Lu, Jiwen},
  booktitle={Proceedings of the Computer Vision and Pattern Recognition Conference},
  pages={19057--19066},
  year={2025}
}

@article{liu2023visual,
  title={Visual instruction tuning},
  author={Liu, Haotian and Li, Chunyuan and Wu, Qingyang and Lee, Yong Jae},
  journal={Advances in neural information processing systems},
  volume={36},
  pages={34892--34916},
  year={2023}
}

@inproceedings{lin2024vila,
  title={Vila: On pre-training for visual language models},
  author={Lin, Ji and Yin, Hongxu and Ping, Wei and Molchanov, Pavlo and Shoeybi, Mohammad and Han, Song},
  booktitle={Proceedings of the IEEE/CVF conference on computer vision and pattern recognition},
  pages={26689--26699},
  year={2024}
}

@inproceedings{li2024llama,
  title={Llama-vid: An image is worth 2 tokens in large language models},
  author={Li, Yanwei and Wang, Chengyao and Jia, Jiaya},
  booktitle={European Conference on Computer Vision},
  pages={323--340},
  year={2024},
  organization={Springer}
}

@article{wei2025streamvln,
  title={Streamvln: Streaming vision-and-language navigation via slowfast context modeling},
  author={Wei, Meng and Wan, Chenyang and Yu, Xiqian and Wang, Tai and Yang, Yuqiang and Mao, Xiaohan and Zhu, Chenming and Cai, Wenzhe and Wang, Hanqing and Chen, Yilun and others},
  journal={arXiv preprint arXiv:2507.05240},
  year={2025}
}

@article{li2025compassnav,
  title={Compassnav: Steering from path imitation to decision understanding in navigation},
  author={Li, LinFeng and Zhao, Jian and Xie, Yuan and Tan, Xin and Li, Xuelong},
  journal={arXiv preprint arXiv:2510.10154},
  year={2025}
}

@article{yang2025qwen3,
  title={Qwen3 technical report},
  author={Yang, An and Li, Anfeng and Yang, Baosong and Zhang, Beichen and Hui, Binyuan and Zheng, Bo and Yu, Bowen and Gao, Chang and Huang, Chengen and Lv, Chenxu and others},
  journal={arXiv preprint arXiv:2505.09388},
  year={2025}
}

@article{bai2025qwen2,
  title={Qwen2. 5-vl technical report},
  author={Bai, Shuai and Chen, Keqin and Liu, Xuejing and Wang, Jialin and Ge, Wenbin and Song, Sibo and Dang, Kai and Wang, Peng and Wang, Shijie and Tang, Jun and others},
  journal={arXiv preprint arXiv:2502.13923},
  year={2025}
}

@inproceedings{krantz2020beyond,
  title={Beyond the nav-graph: Vision-and-language navigation in continuous environments},
  author={Krantz, Jacob and Wijmans, Erik and Majumdar, Arjun and Batra, Dhruv and Lee, Stefan},
  booktitle={European Conference on Computer Vision},
  pages={104--120},
  year={2020},
  organization={Springer}
}

@article{ku2020room,
  title={Room-across-room: Multilingual vision-and-language navigation with dense spatiotemporal grounding},
  author={Ku, Alexander and Anderson, Peter and Patel, Roma and Ie, Eugene and Baldridge, Jason},
  journal={arXiv preprint arXiv:2010.07954},
  year={2020}
}

@article{cheng2024navila,
  title={Navila: Legged robot vision-language-action model for navigation},
  author={Cheng, An-Chieh and Ji, Yandong and Yang, Zhaojing and Gongye, Zaitian and Zou, Xueyan and Kautz, Jan and B{\i}y{\i}k, Erdem and Yin, Hongxu and Liu, Sifei and Wang, Xiaolong},
  journal={arXiv preprint arXiv:2412.04453},
  year={2024}
}

@inproceedings{lin2023learning,
  title={Learning vision-and-language navigation from youtube videos},
  author={Lin, Kunyang and Chen, Peihao and Huang, Diwei and Li, Thomas H and Tan, Mingkui and Gan, Chuang},
  booktitle={Proceedings of the IEEE/CVF International Conference on Computer Vision},
  pages={8317--8326},
  year={2023}
}

@article{tan2019learning,
  title={Learning to navigate unseen environments: Back translation with environmental dropout},
  author={Tan, Hao and Yu, Licheng and Bansal, Mohit},
  journal={arXiv preprint arXiv:1904.04195},
  year={2019}
}

@inproceedings{azuma2022scanqa,
  title={Scanqa: 3d question answering for spatial scene understanding},
  author={Azuma, Daichi and Miyanishi, Taiki and Kurita, Shuhei and Kawanabe, Motoaki},
  booktitle={proceedings of the IEEE/CVF conference on computer vision and pattern recognition},
  pages={19129--19139},
  year={2022}
}

@inproceedings{chen2024sharegpt4v,
  title={Sharegpt4v: Improving large multi-modal models with better captions},
  author={Chen, Lin and Li, Jinsong and Dong, Xiaoyi and Zhang, Pan and He, Conghui and Wang, Jiaqi and Zhao, Feng and Lin, Dahua},
  booktitle={European Conference on Computer Vision},
  pages={370--387},
  year={2024},
  organization={Springer}
}

@inproceedings{maaz2024video,
  title={Video-chatgpt: Towards detailed video understanding via large vision and language models},
  author={Maaz, Muhammad and Rasheed, Hanoona and Khan, Salman and Khan, Fahad},
  booktitle={Proceedings of the 62nd Annual Meeting of the Association for Computational Linguistics (Volume 1: Long Papers)},
  pages={12585--12602},
  year={2024}
}

@inproceedings{radford2021learning,
  title={Learning transferable visual models from natural language supervision},
  author={Radford, Alec and Kim, Jong Wook and Hallacy, Chris and Ramesh, Aditya and Goh, Gabriel and Agarwal, Sandhini and Sastry, Girish and Askell, Amanda and Mishkin, Pamela and Clark, Jack and others},
  booktitle={International conference on machine learning},
  pages={8748--8763},
  year={2021},
  organization={PmLR}
}

@article{kahneman2011fast,
  title={Fast and slow thinking},
  author={Kahneman, Daniel},
  journal={Allen Lane and Penguin Books, New York},
  year={2011}
}

@article{feng2025image,
  title={Image-Goal Navigation Using Refined Feature Guidance and Scene Graph Enhancement},
  author={Feng, Zhicheng and Chen, Xieyuanli and Shi, Chenghao and Luo, Lun and Chen, Zhichao and Liu, Yun-Hui and Lu, Huimin},
  journal={arXiv preprint arXiv:2503.10986},
  year={2025}
}

@article{qin2025navigatediff,
  title={Navigatediff: Visual predictors are zero-shot navigation assistants},
  author={Qin, Yiran and Sun, Ao and Hong, Yuze and Wang, Benyou and Zhang, Ruimao},
  journal={arXiv preprint arXiv:2502.13894},
  year={2025}
}

@inproceedings{zhai2023sigmoid,
  title={Sigmoid loss for language image pre-training},
  author={Zhai, Xiaohua and Mustafa, Basil and Kolesnikov, Alexander and Beyer, Lucas},
  booktitle={Proceedings of the IEEE/CVF international conference on computer vision},
  pages={11975--11986},
  year={2023}
}

@article{dubey2024llama,
  title={The llama 3 herd of models},
  author={Dubey, Abhimanyu and Jauhri, Abhinav and Pandey, Abhinav and Kadian, Abhishek and Al-Dahle, Ahmad and Letman, Aiesha and Mathur, Akhil and Schelten, Alan and Yang, Amy and Fan, Angela and others},
  journal={arXiv preprint arXiv:2407.21783},
  year={2024}
}

@inproceedings{papineni2002bleu,
  title={Bleu: a method for automatic evaluation of machine translation},
  author={Papineni, Kishore and Roukos, Salim and Ward, Todd and Zhu, Wei-Jing},
  booktitle={Proceedings of the 40th annual meeting of the Association for Computational Linguistics},
  pages={311--318},
  year={2002}
}

@inproceedings{banerjee2005meteor,
  title={METEOR: An automatic metric for MT evaluation with improved correlation with human judgments},
  author={Banerjee, Satanjeev and Lavie, Alon},
  booktitle={Proceedings of the acl workshop on intrinsic and extrinsic evaluation measures for machine translation and/or summarization},
  pages={65--72},
  year={2005}
}

@inproceedings{vedantam2015cider,
  title={Cider: Consensus-based image description evaluation},
  author={Vedantam, Ramakrishna and Lawrence Zitnick, C and Parikh, Devi},
  booktitle={Proceedings of the IEEE conference on computer vision and pattern recognition},
  pages={4566--4575},
  year={2015}
}

@inproceedings{lin2004rouge,
  title={Rouge: A package for automatic evaluation of summaries},
  author={Lin, Chin-Yew},
  booktitle={Text summarization branches out},
  pages={74--81},
  year={2004}
}

@article{wu2024cr,
  title={CR-former: Single-image cloud removal with focused Taylor attention},
  author={Wu, Yang and Deng, Ye and Zhou, Sanping and Liu, Yuhan and Huang, Wenli and Wang, Jinjun},
  journal={IEEE Transactions on Geoscience and Remote Sensing},
  volume={62},
  pages={1--14},
  year={2024},
  publisher={IEEE}
}

@inproceedings{wu2025event,
  title={Event-Equalized Dense Video Captioning},
  author={Wu, Kangyi and Li, Pengna and Fu, Jingwen and Li, Yizhe and Wu, Yang and Liu, Yuhan and Wang, Jinjun and Zhou, Sanping},
  booktitle={Proceedings of the IEEE/CVF Conference on Computer Vision and Pattern Recognition},
  pages={8417--8427},
  year={2025}
}

@inproceedings{liu2024semantic,
  title={Semantic-aware representation learning for homography estimation},
  author={Liu, Yuhan and Huang, Qianxin and Hui, Siqi and Fu, Jingwen and Zhou, Sanping and Wu, Kangyi and Li, Pengna and Wang, Jinjun},
  booktitle={Proceedings of the 32nd ACM International Conference on Multimedia},
  pages={2506--2514},
  year={2024}
}

@inproceedings{liu2025mind,
  title={Mind the gap: Aligning vision foundation models to image feature matching},
  author={Liu, Yuhan and Fu, Jingwen and Wu, Yang and Wu, Kangyi and Li, Pengna and Wu, Jiayi and Zhou, Sanping and Xin, Jingmin},
  booktitle={Proceedings of the IEEE/CVF International Conference on Computer Vision},
  pages={20313--20323},
  year={2025}
}

@article{qi2026patchcue,
  title={PatchCue: Enhancing Vision-Language Model Reasoning with Patch-Based Visual Cues},
  author={Qi, Yukun and Fu, Pei and Li, Hang and Liu, Yuhan and Jiang, Chao and Qin, Bin and Luo, Zhenbo and Luan, Jian},
  journal={arXiv preprint arXiv:2603.05869},
  year={2026}
}

@article{fu2024understanding,
  title={Understanding mobile GUI: From pixel-words to screen-sentences},
  author={Fu, Jingwen and Zhang, Xiaoyi and Wang, Yuwang and Zeng, Wenjun and Zheng, Nanning},
  journal={Neurocomputing},
  volume={601},
  pages={128200},
  year={2024},
  publisher={Elsevier}
}

@article{wu2025att,
  title={ATT-CR: Adaptive Triangular Transformer for Cloud Removal},
  author={Wu, Yang and Deng, Ye and Li, Pengna and Huang, Wenli and Wu, Kangyi and Xin, Xiaomeng and Wang, Jinjun},
  journal={IEEE Journal of Selected Topics in Applied Earth Observations and Remote Sensing},
  year={2025},
  publisher={IEEE}
}

@article{lyu2026himemvln,
  title={HiMemVLN: Enhancing Reliability of Open-Source Zero-Shot Vision-and-Language Navigation with Hierarchical Memory System},
  author={Lyu, Kailin and Wu, Kangyi and Li, Pengna and Hu, Xiuyu and Si, Qingyi and Miao, Cui and Yang, Ning and Wang, Zihang and Xiao, Long and Hu, Lianyu and others},
  journal={arXiv preprint arXiv:2603.14807},
  year={2026}
}

@article{li2024camera,
  title={Camera-aware label refinement for unsupervised person re-identification},
  author={Li, Pengna and Wu, Kangyi and Huang, Wenli and Zhou, Sanping and Wang, Jinjun},
  journal={arXiv preprint arXiv:2403.16450},
  year={2024}
}

@misc{wu2025cemnetcrossemotionmemorynetwork,
      title={CEM-Net: Cross-Emotion Memory Network for Emotional Talking Face Generation}, 
      author={Kangyi Wu and Pengna Li and Jingwen Fu and Yang Wu and Yuhan Liu and Sanping Zhou and Jinjun Wang},
      year={2025},
      eprint={2508.12368},
      archivePrefix={arXiv},
      primaryClass={cs.MM},
      url={https://arxiv.org/abs/2508.12368}, 
}

\appendix
\newpage
\section*{Appendix}
\label{sec:appendix}

This supplementary material provides additional details on the proposed method and experimental results that could not be included in the main manuscript due to page limitations.

Specifically, the structure of this appendix is organized as follows:

\textbf{Section~\ref{potential risk}} discusses the potential risk and data consent in the checklist.

\textbf{Section~\ref{app:model}} provides more method details, including the high-level/low-level modules, navigation policy, prompts for dataset construction, and the wandering suppression penalty design.

\textbf{Section~\ref{app:exp}} summarizes experimental settings, including evaluation metrics, datasets, implementation details and data quality control details.

\textbf{Section~\ref{app:results}} reports additional results omitted from the main paper, including fine-grained cross-domain evaluation, reward-weight ablations, efficiency analysis and more qualitative visualizations.

\begin{figure*}[t]
  \centering
  \includegraphics[width=\linewidth]{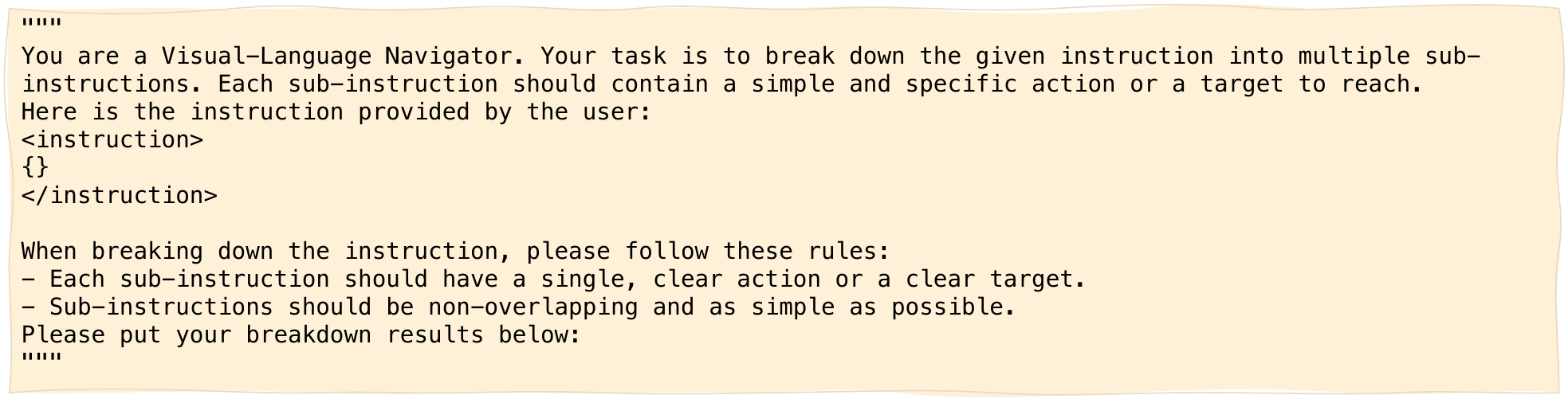}
  \caption{Sub-task decomposition prompt.
    We prompt the LLM to break down a long navigation instruction into a sequence of atomic sub-instructions.}
  \label{fig:prompt_sub_task}
\end{figure*}


\section{Checklist}
\label{potential risk}
\subsection{Potential risk}
Our work targets image-goal navigation in indoor environments and is primarily evaluated in simulation, so the overall risk profile is limited.
The main practical risk is that behaviors learned in simulation may not fully transfer to real robots due to differences in sensing and embodiment, which could lead to occasional navigation inefficiency or minor unintended contacts in cluttered spaces.
In addition, the high-level vision-language reasoning module may inherit typical robustness limitations of large models, resulting in degraded performance under rare visual conditions.
These issues are not specific to our method and can be mitigated with cautious deployment (e.g., safe-speed operation, collision avoidance layers) and improved sim-to-real modeling in future work.

\subsection{Data Contains Personally Identifying Info Or Offensive Content}

Our experiments primarily use standard simulation datasets (Gibson, MP3D, HM3D), which contain reconstructed indoor environments and do not include personal identifiers such as names, faces, voices, or other uniquely identifying information. For our real-world demonstrations, we collected only egocentric RGB observations of indoor scenes in a controlled setting.

\subsection{Data Consent}
We use only publicly released datasets that are obtained through the official application/registration process and downloaded under the corresponding licenses/terms of use. We did not collect or curate new data from identifiable individuals. Consent for any original data collection is governed by the dataset providers, and our usage follows the permitted research use terms specified by the official data access agreements. 

\subsection{Information About Use Of AI Assistants}
We used an AI assistant to support scientific writing and language polishing (e.g., improving clarity, grammar, and formatting). All technical content, experimental design, results, and claims were produced and verified by the authors, who also reviewed and edited the final manuscript to ensure correctness and originality.

\section{More Method Details}
\label{app:model}
\subsection{High-level reasoning module}
We follow the model design of NaVILA~\cite{cheng2024navila}: a vision encoder~(SigLIP~\cite{zhai2023sigmoid}) extracts visual tokens from the history frames, current observation, and goal image; an MLP projector maps the visual tokens into the language embedding space; and an LLM~(LLaMA3~\cite{dubey2024llama}) policy head performs auto-regressive decoding to predict the next sub-task instruction.

\subsection{Low-level prediction module}
For the semantic sub-task channel, we encode the predicted sub-task instruction, the current observation, and the goal image using CLIP text/vision encoders (ResNet50 backbone)~\cite{he2016deep, radford2021learning,li2024camera}.
For the navigation-specific channel, we concatenate the current observation and the goal image along the channel dimension and extract navigation features. The navigation encoder is initialized with a ResNet50~\cite{he2016deep} pretrained in Imagenet~\cite{deng2009imagenet}.
We fuse the semantic features and navigation features via a 2-layer MLP fusion module before feeding them to the navigation policy.

\begin{figure*}[t]
    \centering
    \includegraphics[width=\linewidth]{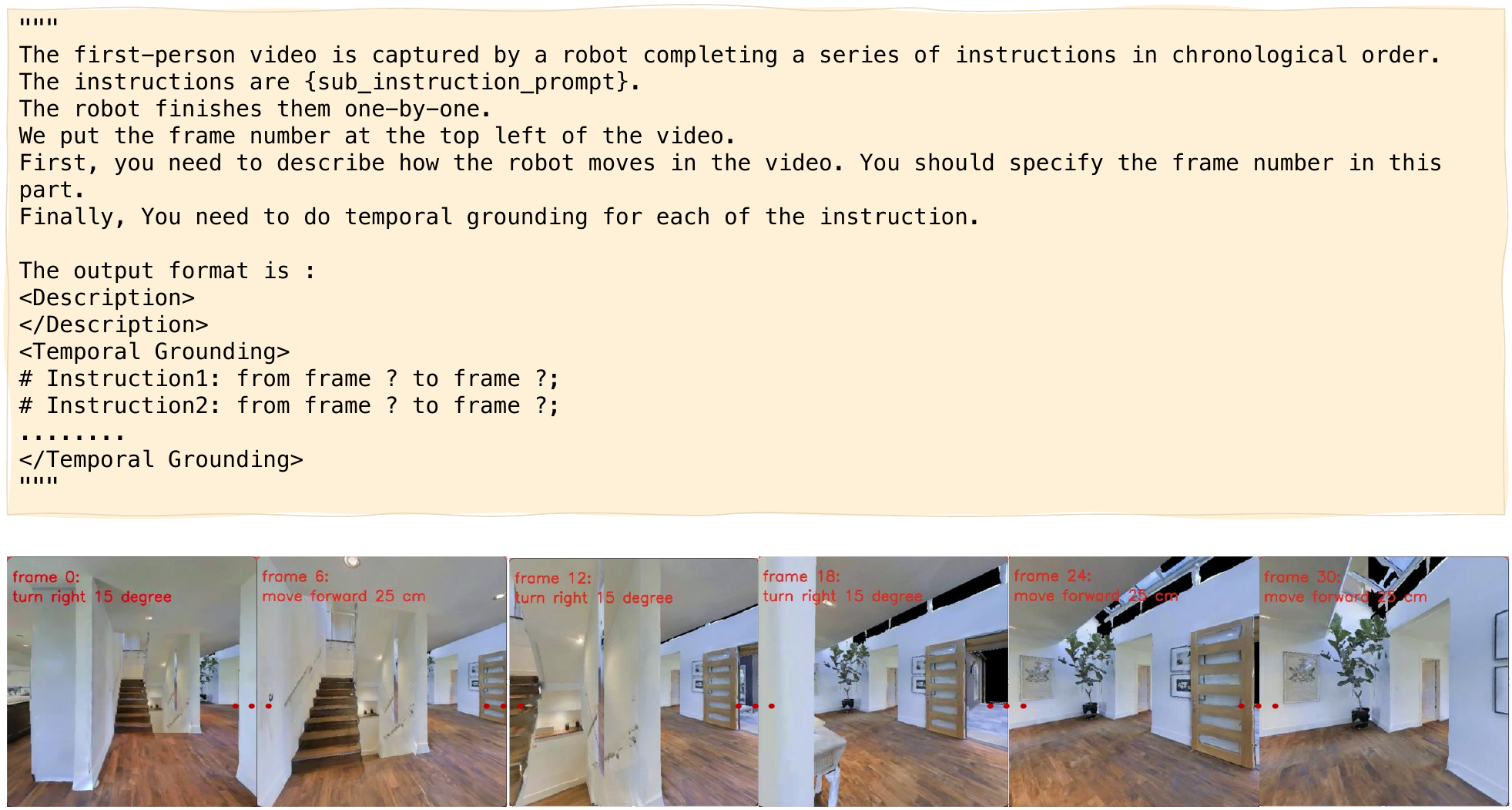}
    \caption{Temporal grounding prompt and annotated video clip.
    \textbf{Top:} The prompt instructs the model to assign a temporal interval to each sub-instruction using a structured output format.
    \textbf{Bottom:} an example input video clip used for grounding.
    To facilitate temporal reasoning, we overlay the \emph{frame number} and the \emph{executed action} at the top-left corner of each frame.}
    \label{fig:temporal_grounding_clip}
\end{figure*}

\subsection{Navigation policy}
\label{navigation policy}
The navigation policy is composed of a 2-layer GRU with a 128-dimensional embedding size.
We train the navigation policy using reinforcement learning.
To enhance the generalization ability of the agent, we follow previous reinforcement learning methods~\cite{mezghan2022memory,al2022zero,wu2024cr} to apply data augmentation to the visual observation and goal images.
We adopt two kinds of augmentations: (1) random cropping, which enlarges the input image and then selects a random portion of the original size; (2) colour jitter, which randomly adjusts the brightness, contrast, saturation, and hue of the image.
During navigation, the agent obtains the current observation from the sensors at each step.
We sequentially apply both data augmentation methods to these images, which can enhance the visual diversity in the training data.

After the data augmentation, the visual observation and goal images are fed into the semantic and navigation channel to obtain fused features. The agent receives the fused features $f_{fused}$ and passes them into the navigation policy $\pi$.
The policy utilizes the fused features along with a representation of previous actions to predict the agent's current state embedding $s_t$ at time $t$, and we formulate this as follows:
\begin{equation}
s_t = \pi(f_{fused} \oplus a_{t-1} \mid h_{t-1}),
\end{equation}
where $a_{t-1}$ denotes the representation of previous actions and $h_{t-1}$ is the hidden layer of the recurrent layers in the policy from the previous step.
An actor-critic network then utilizes $s_t$ to predict the state value and determine the agent’s next action.

\subsection{Prompts for Dataset Construction} For training the high-level reasoning module, we construct a hierarchical reasoning dataset from existing VLN trajectory data. As described in Sec.~\ref{sec:Hierarchical Reasoning Dataset}, we prompt Qwen3~\cite{yang2025qwen3} to decomposes a long-horizon instruction into a sequence of non-overlapping, atomic sub-instructions, where each sub-instruction contains a single clear action or a specific target to reach. Then we prompt visual language model Qwen2-VL~\cite{bai2025qwen2} to performs temporal grounding by analyzing the changes of first-person observations over time and selecting which sub-instruction is currently being carried out.
For readability, we optionally visualize these prompts and the corresponding input/output format.

\paragraph{Sub-task decomposition prompt}
As illustrated in Fig.~\ref{fig:prompt_sub_task}, we prompt the LLM to break down the given navigation instruction into multiple sub-instructions with clear and minimal objectives, ensuring each sub-instruction is simple and does not overlap with others.

\paragraph{Temporal grounding prompt.} The prompt is shown in Fig.~\ref{fig:temporal_grounding_clip}. 
Given the ordered sub-instructions and a first-person video clip from time $0$ to $t$~(each video frame is augmented with the frame number and the action executed at that time step.), we prompt the VLM to infer which sub-instruction the agent is executing based on \emph{temporal changes} in the observed frames, rather than static scene recognition.
This grounding step produces an alignment between sub-instructions and temporal segments.

\subsection{ZER Reward}\label{zer reward}
\paragraph{ZER Reward.}Following ZER~\cite{al2022zero}, we use a dense distance-and-view shaping reward together with a sparse success reward.
After taking action $a_t$, the step reward is
\begin{equation}
r_t = (d_{t-1}-d_t) + \mathbb{I}(d_t\le d_s)(\alpha_{t-1}-\alpha_t) - \gamma 
\label{eq:reward_zer}
\end{equation}
where $d_t$ is the geodesic distance to the goal, $\alpha_t$ is the minimum angular difference (in radians) between the agent view and the goal views and $\gamma$ is the slack reward for efficiency.
The view term is activated only within the success area ($d_t\le d_s$, where $d_s$ is the success distance).
The sparse success reward is
\begin{equation}
R_s = 5\Big[\mathbb{I}(d_t\le d_s) + \mathbb{I}(d_t\le d_s \wedge \alpha_t\le \alpha_s)\Big],
\label{eq:success_zer}
\end{equation}
with $d_s=1$m and $\alpha_s=25^\circ$. The agent receives a sparse reward when it reaches the success area and stops within an view angle $\alpha_s$ from the goal angle.

Although the ZER reward provides dense shaping, its distance term $(d_{t-1}-d_t)$ is local and does not explicitly penalize redundant motions: when the agent explores ambiguous corridors or makes wrong turns, it can easily fall into short-range oscillations (e.g., left-right rotations or backtracking), where the net progress in geodesic distance becomes small and noisy.

\subsection{Wandering Suppression Penalty Details}
\label{wsp}
\paragraph{Path-length penalty.}
The Path-length penalty penalize unnecessarily long trajectories by incorporating the cumulative traveled length into the potential.
Let $\mathbf{p}_t\in\mathbb{R}^3$ be the agent position at time step $t$.
We define the step displacement as the Euclidean distance between two consecutive positions:
\begin{equation}
\Delta \ell_t = \|\mathbf{p}_t - \mathbf{p}_{t-1}\|_2,
\label{eq:step_len}
\end{equation}
and accumulate it over time to obtain the traveled path length:
\begin{equation}
\ell_t = \sum_{\tau=1}^{t}\Delta \ell_\tau,\qquad \ell_0=0.
\label{eq:path_len}
\end{equation}
This term is monotonic increasing and thus yields a consistent penalty.

\noindent\textbf{Revisit penalty.}
In this part, we provides more design details about the revisit penalty.
At each time step $t$, let $\mathbf{p}_t\in\mathbb{R}^3$ denote the agent position.
We discretize $\mathbf{p}_t$ into a voxel index (grid key) $\kappa_t$ with resolution $s$ by
\begin{equation}
\kappa_t = \big(\lfloor p_x/s \rfloor,\ \lfloor p_y/s \rfloor,\ \lfloor p_z/s \rfloor\big),
\label{eq:quantize}
\end{equation}
where $s$ corresponds to \texttt{revisit\_radius} in our implementation and it is set to 0.25.
We maintain a visited set $\mathcal{V}$ that stores voxel keys observed so far in the episode.
The revisit cost is then computed as
\begin{equation}
c_t = \lambda_{\text{rv}}\,\mathbb{I}(\kappa_t \in \mathcal{V}),
\qquad
\mathcal{V} \leftarrow \mathcal{V} \cup \{\kappa_t\},
\label{eq:revisit_cost}
\end{equation}
where $\lambda_{\text{rv}}$ corresponds to \texttt{revisit\_weight}, which is set to 0.02.
Intuitively, whenever the agent returns to a previously visited voxel, it receives an additional penalty, discouraging redundant backtracking and local dithering.

Finally, the path length and revisit cost are integrated into the shaping function as
\begin{equation}
\Phi_t = \lambda_w \big(\ell_t + c_t\big) + d_t + \alpha_t,
\label{eq:potential_with_revisit}
\end{equation}
where $\ell_t$ is the cumulative traveled length, $c_t$ is the reivisit cost, $d_t$ is the geodesic distance to the goal, and $\alpha_t$ is the view-angle distance to the closest goal view. $\lambda_w$ is the hyperparameter weight. 
The step reward is obtained by the potential difference, i.e., $r_t \leftarrow r_t + (\Phi_{t-1}-\Phi_t)$, so that our new design reward consistently decrease the reward and thus suppress wandering.

\paragraph{Using Wandering Suppression Penalty (WSP).}
While WSP provides consistent negative feedback for redundant motions, directly enabling it from scratch may harm early-stage exploration.
In the initial phase of PPO training, the policy is largely uninformative and relies on stochastic exploration to discover trajectories that reduce $d_t$ and eventually enter the success area.
If WSP is applied too early, the additional penalties on traveled length and revisits can dominate the learning signal, making ``doing nothing'' (or minimal motion) a locally attractive behavior since it avoids accumulating $\ell_t$ and $c_t$.
This may lead to overly conservative policies and slow down (or even stall) learning.

To mitigate this issue, we adopt a two-stage training schedule.
We first pretrain the policy using the original ZER reward (Eqs.~\eqref{eq:reward_zer}--\eqref{eq:success_zer}) until the navigation behavior becomes reasonably stable (i.e., the agent consistently makes progress toward the goal).
Then, we activate WSP and continue training with the combined reward.
In this second stage, the policy already has a meaningful notion of goal-directed movement, and WSP mainly acts as a regularizer that suppresses dithering and detours, thereby reducing wandering and improving trajectory efficiency.

In practice, we find that WSP takes effect quickly once activated.
After the initial warm-up with the ZER reward, enabling WSP and continuing training for about 10M environment steps is sufficient to noticeably suppress wandering behaviors.

\begin{table*}[t]
\centering
\small
\begin{tabular}{lcccc}
\toprule
Data & \# Samples & Format & Temporal & Semantic \\
 &  & Compliance & Consistency & Grounding \\
\midrule
Before TQCM & 1328K & 93.2\% & 87.2\% & 84.2\% \\
After TQCM & 767K & 99.9\% & 99.6\% & 98.6\% \\
\bottomrule
\end{tabular}
\caption{Quality evaluation of the generated Hierarchical Reasoning Dataset. TQCM substantially improves format compliance, temporal consistency, and semantic grounding accuracy.}
\label{tab:dataset_quality}
\end{table*}
\section{Experimental Details}
\label{app:exp}

\subsection{Evaluation metric details}
We evaluate image-goal navigation performance with the success rate~(SR) and success weighted by path length~(SPL)~\cite{anderson2018evaluation}.

\noindent\textbf{SR~(Success Rate)}:
SR evaluates the success rate of the agent in successfully navigating to the target location.
The maximum success distance is 1m from the real target location.
SR is defined by:
\begin{equation}
SR = \frac{1}{{N_e}} \sum_{i=1}^{N_e} S_i,
\end{equation}
\begin{equation}
S_i =
\left\{
\begin{aligned}
1,&~\mathrm{if~the~}i\mathrm{th~episode~succeeds},\\
0,&~\mathrm{if~the~}i\mathrm{th~episode~fails},
\end{aligned}
\right.
\end{equation}
where $N_e$ represents the total number of test episodes and $S_i$ indicates whether the $i$th episode is successful.

\noindent\textbf{SPL~(Success weighted by Path Length)}:
SPL considers the navigation path length along with the success rate:
\begin{equation}
SPL = \frac{1}{{N_e}} \sum_{i=1}^{N_e} S_i \frac{l_i}{\max(p_i,l_i)},
\end{equation}
where $l_i$ is the shortest path distance from the agent's starting position to the goal location in the $i$th episode and $p_i$ is the predicted navigation path distance executed by the agent.

\subsection{Dataset details}
\label{sec2}
We trained REGNav in the Gibson dataset~\cite{xiazamirhe2018gibsonenv} and inference in the Gibson, Matterport3D~(MP3D)~\cite{chang2017matterport3d} and HM3D~\cite{ramakrishnan2021habitat} datasets.
Specifically, Gibson is composed of full indoor environments, which have multi-rooms.
The training set in Gibson comprises 9000 episodes sampled from 72 Gibson training scenes.
These episodes are evenly distributed across three difficulty levels based on the goal location's geodesic distance from the starting location:
easy (1.5--3m), medium (3--5m), and hard (5--10m).
The test set contains 4200 episodes sampled from 14 unseen scenes and covers the same three difficulty levels.
The test scenes are separate from the training scenes to evaluate the agent's ability to generalize to previously unseen environments.
Compared with Gibson, MP3D comprises more complex and expansive scenes and HM3D provides more diverse scenes.
For evaluation on MP3D and HM3D, we adopt the same test splits as ZER~\cite{al2022zero}.
The test set of MP3D includes 1000 episodes of 100 scenes per difficulty level while HM3D contains 1000 episodes of 18 scenes per level.

\subsection{Inplementation Details}
For the high-level module training stage, we use a learning rate of 1e-4 with cosine decay and a warm-up ratio of 0.03. 
The low-level modules are trained end-to-end using the Adam optimizer~\cite{kingma2014adam} with DD-PPO~\cite{wijmans2019dd}, using 64 forward steps, an entropy coefficient of 0.01, and a clipping value of 0.2. 
The slack reward $\gamma$ is set to -0.01. In the simulation environments, the height of agent is set to 1.5m and the radius is 0.1m. The agent has a single RGB sensor with a $90^{\circ}$ FOV and 128×128 resolution. The action space consists of MOVE\_FORWARD by 0.25m, TURN\_LEFT, TURN\_RIGHT by $30^{\circ}$ and STOP.
Our HRNav is trained on a server with 8 NAVIDIA H20 GPUs. The high-level module finetuning requires 64 hours, and the low-level module requires 30 hours. During inference, we run the low-level policy for 15 steps, and then invoke the high-level module once.
We will release both our training code and data upon paper publication.

\begin{table*}[t]
\centering
\setlength{\tabcolsep}{8pt}
\renewcommand\arraystretch{1.3}
\begin{tabular}{c|cc|cc|cc|cc}
\hline
\multirow{3}{*}{\textbf{Method}}
&\multicolumn{8}{|c}{\textbf{MP3D}}\\
\cline{2-9}
&\multicolumn{2}{|c|}{\textbf{Easy}}&\multicolumn{2}{|c|}{\textbf{Medium}}&\multicolumn{2}{|c}{\textbf{Hard}}&\multicolumn{2}{|c}{\textbf{Overall}}\\
\cline{2-9}
&SPL$\uparrow$&SR$\uparrow$&SPL$\uparrow$&SR$\uparrow$&SPL$\uparrow$&SR$\uparrow$&SPL$\uparrow$&SR$\uparrow$\\
\hline
FGPrompt-MF&56.4\%&88.1\%&44.6\%&77.8\%&32.0\%&60.1\%&44.3\%&75.3\%\\
FGPrompt-EF&{61.6\%}&{88.9\%}&{50.6\%}&{78.7\%}&{34.1\%}&59.5\%&{48.8\%}&{75.7\%}\\
REGNav&\underline{61.7\%}&\underline{90.0\%}&\underline{52.0\%}&\underline{81.4\%}&\underline{36.9\%}&\underline{62.7\%}&\underline{50.2\%}&\underline{78.0\%}\\
HRNav&\textbf{64.6\%}&\textbf{90.9\%}&\textbf{58.3\%}&\textbf{84.2\%}&\textbf{45.9\%}&\textbf{69.1\%}&\textbf{56.3\%}&\textbf{81.1\%}\\
\hline
\end{tabular}
\caption{\textbf{Fine-grained cross-domain evaluation on MP3D.}
We compare HRNav against FGPrompt~\cite{sun2024fgprompt} and REGNav~\cite{li2025regnav} across three difficulty levels on MP3D~\cite{chang2017matterport3d}.
Each difficulty level contains 1000 episodes sampled from 100 unseen scenes.
All methods are trained on Gibson~\cite{xiazamirhe2018gibsonenv} and directly transferred to MP3D for zero-shot inference.
The \textbf{best} and \underline{second-best} results are highlighted.}
\label{tab:sota-mp3d}
\end{table*}

\subsection{Dataset Quality Control}
\label{app:dataset_quality}

Since the Hierarchical Reasoning Dataset is automatically constructed with LLMs and VLMs, we further conduct a quality-control evaluation to assess the reliability of the generated temporal grounding and sub-task annotations. We consider three complementary aspects: format compliance, temporal consistency, and semantic grounding accuracy.

\paragraph{Format Compliance.}
We first evaluate whether the generated temporal grounding annotations follow the required structured format. Each annotation line is expected to specify the sub-task index and its corresponding temporal interval, e.g.,
\emph{\# Instruction1: from frame 0 to frame 8;} or \emph{\# Instruction1: from frame 0 onwards;}. 
An annotation is considered format-compliant if all its lines can be successfully parsed into a valid tuple consisting of the instruction index, start frame, and end frame or onwards marker. In addition, the instruction indices are required to be continuous and start from 1, without missing or duplicated indices. This metric reflects whether the generated annotations can be reliably used for downstream dataset construction.

\paragraph{Temporal Consistency.}
We then check whether the grounded sub-tasks are temporally consistent with the navigation trajectory. After sorting the parsed intervals by their instruction indices, we require the start frames of consecutive sub-tasks to be strictly increasing. We also require that adjacent intervals do not overlap under the end-exclusive interval convention, i.e., an interval from frame $s$ to frame $e$ is interpreted as $[s,e)$. For annotations with an \emph{onwards} marker, the interval is extended to the last available frame of the trajectory. An annotation is considered temporally consistent only if all sub-task intervals satisfy the above ordering and non-overlap constraints. This metric evaluates whether the decomposed sub-tasks form a valid temporal progression along the trajectory.

\paragraph{Semantic Grounding Accuracy.}
Finally, we evaluate whether each grounded sub-task is semantically supported by the visual observations within its assigned temporal interval. For each sub-task interval, we uniformly sample several frames from the interval and provide them, together with the corresponding sub-task instruction, to a VLM judge. The judge is prompted to determine whether the instruction is visually supported by the sampled frames and to return a structured JSON output containing a binary consistency decision, evidence frames, confidence score, and a short reason. We report semantic grounding accuracy as the percentage of judged sub-task intervals that are classified as semantically consistent. This metric measures whether the temporal interval not only satisfies structural constraints but also visually corresponds to the intended navigation sub-task.

\paragraph{Analysis.}
As shown in Table~\ref{tab:dataset_quality}, the raw automatically generated annotations already achieve reasonable quality, but still contain non-negligible noise. Before quality control, 93.2\% of the annotations satisfy the required output format, while 87.2\% pass the temporal consistency check, indicating that some generated sub-task intervals contain malformed structures, missing or duplicated indices, or overlapping temporal boundaries. The semantic grounding accuracy is 84.2\%, suggesting that a portion of the automatically grounded intervals do not fully match the visual evidence in the corresponding frames.

After applying the proposed Triple Quality Control Mechanism (TQCM), the annotation quality is significantly improved. The filtered dataset achieves 99.9\% format compliance and 99.6\% temporal consistency, showing that the remaining annotations are structurally valid and temporally well ordered. The semantic grounding accuracy also increases to 98.6\%, indicating that most retained sub-task annotations are visually supported by their grounded frame intervals. Although this filtering process reduces the dataset size from 1328K to 767K samples, it removes noisy annotations and yields a cleaner training corpus for high-level planning. This quality-control process improves the reliability and reproducibility of the constructed dataset, which is important for training the VLM-based slow planner.

\begin{table*}[t]
\centering
\setlength{\tabcolsep}{8pt}
\renewcommand\arraystretch{1.3}
\begin{tabular}{c|cc|cc|cc|cc}
\hline
\multirow{3}{*}{\textbf{Method}}
&\multicolumn{8}{|c}{\textbf{HM3D}}\\
\cline{2-9}
&\multicolumn{2}{|c|}{\textbf{Easy}}&\multicolumn{2}{|c|}{\textbf{Medium}}&\multicolumn{2}{|c}{\textbf{Hard}}&\multicolumn{2}{|c}{\textbf{Overall}}\\
\cline{2-9}
&SPL$\uparrow$&SR$\uparrow$&SPL$\uparrow$&SR$\uparrow$&SPL$\uparrow$&SR$\uparrow$&SPL$\uparrow$&SR$\uparrow$\\
\hline
FGPrompt-MF&44.0\%&83.2\%&40.8\%&75.6\%&31.8\%&62.6\%&38.9\%&73.8\%\\
FGPrompt-EF&{47.3\%}&{85.0\%}&{44.2\%}&{77.8\%}&\underline{34.9\%}&\underline{62.8\%}&{42.1\%}&{75.2\%}\\
REGNav&\underline{50.6\%}&\underline{85.4\%}&\underline{47.0\%}&\underline{78.9\%}&{34.3\%}&61.3\%&\underline{44.0\%}&\underline{75.2\%}\\
HRNav&\textbf{54.3\%}&\textbf{89.6\%}&\textbf{52.3\%}&\textbf{83.8\%}&\textbf{41.3\%}&\textbf{66.5\%}&\textbf{49.3\%}&\textbf{80.0\%}\\
\hline
\end{tabular}
\caption{\textbf{Fine-grained cross-domain evaluation on HM3D.}
We compare HRNav against FGPrompt~\cite{sun2024fgprompt} and REGNav~\cite{li2025regnav} across three difficulty levels on HM3D~\cite{szot2021habitat}.
Each difficulty level contains 1000 episodes sampled from 18 unseen scenes.
All methods are trained on Gibson~\cite{xiazamirhe2018gibsonenv} and directly transferred to HM3D for zero-shot evaluation.
The \textbf{best} and \underline{second-best} results are highlighted.}
\label{tab:sota-hm3d}
\end{table*}

\section{Additional Results}
\label{app:results}
We provide additional experimental results that are omitted from the main paper due to space constraints, including:
(i) fine-grained evaluation on MP3D and HM3D (per difficulty level),
(ii) ablation on the weights of the wandering suppression penalty,
and (iii) more qualitative visualizations.

\subsection{Cross-domain evaluation on different difficulty levels.}
To examine the cross-domain generalization of image-goal navigation, we report fine-grained results on MP3D and HM3D by stratifying episodes into three difficulty levels (Easy/Medium/Hard), where the \emph{Hard} split contains the longest trajectories and requires traversing multiple rooms.
As shown in Table~\ref{tab:sota-mp3d} and Table~\ref{tab:sota-hm3d}, all methods are trained on Gibson and directly transferred to unseen scenes for zero-shot evaluation.
We compare HRNav against the FGPrompt baseline~\cite{sun2024fgprompt} and  REGNav~\cite{li2025regnav}.

\paragraph{Analysis.}
HRNav consistently outperforms prior methods across all difficulty levels on both datasets, with the most pronounced gains on \textbf{Hard} trajectories.
On MP3D, HRNav achieves \textbf{45.9\% SPL} and \textbf{69.1\% SR} on the Hard split, surpassing the strongest baseline (REGNav) by \textbf{+9.0 SPL} and \textbf{+6.4 SR}, respectively.
On HM3D, HRNav further improves to \textbf{41.3\% SPL} and \textbf{66.5\% SR} on Hard, yielding \textbf{+6.4 SPL} and \textbf{+3.7 SR} over the best competing method.
These results indicate that the advantage of HRNav becomes larger as the navigation horizon increases: when the goal is far from the start and direct visual matching becomes less reliable, HRNav’s hierarchical reasoning and wandering suppression penalty better reduce wandering and help the agent maintain consistent progress toward the goal, leading to higher success and more efficient paths in the most challenging settings.

\begin{table}[t]
\centering
\setlength{\tabcolsep}{18pt}
\renewcommand\arraystretch{1}
\begin{tabular}{l|cc}
\hline
\textbf{Setting} & \textbf{SR$\uparrow$} & \textbf{SPL$\uparrow$} \\
\hline
1.0 & 71.83\% & 65.86\% \\
0.8 & 81.02\% & 70.75\% \\
0.6 & 85.05\% & 71.91\% \\
0.5 & 89.40\% & \textbf{73.30\%} \\
0.4 & 88.57\% & 72.59\% \\
0.3 & 90.57\% & 71.83\% \\
0.2 & \textbf{94.00\%} & 71.20\% \\
0.1 & 91.88\% & 72.68\% \\
\hline
\end{tabular}
\caption{Ablation study on the weight $\lambda_w$ of wandering suppression penalty.}
\label{tab:ablation_reward_weight}
\end{table}

\begin{table*}[t]
\centering
\small
\begin{tabular}{lcccccc}
\toprule
Step Interval ($k$) & SR (\%) & SPL (\%) & Avg Latency & FPS & Slow-latency & Fast-latency \\
 &  &  & (ms) &  & (ms) & (ms) \\
\midrule
5  & 94.7 & 71.5 & 96.81 & 10.33 & 395.22 & 15.92 \\
10 & 94.4 & 71.1 & 55.08 & 18.13 & 383.28 & 14.45 \\
15 (Ours) & 94.0 & 71.2 & 41.16 & 24.29 & 374.12 & 14.22 \\
30 & 93.4 & 70.6 & 29.75 & 33.61 & 389.47 & 15.08 \\
60 & 92.2 & 69.8 & 24.26 & 41.22 & 408.27 & 15.41 \\
\bottomrule
\end{tabular}
\caption{
Efficiency analysis under different planning intervals. 
The high-level planner is invoked once every $k$ steps. 
The reported average latency per step amortizes the sparse VLM calls over the full navigation trajectory.
}
\label{tab:efficiency}
\end{table*}

\subsection{Ablation study on weights of wandering suppression penalty.}
\label{ablation on weights}
We ablate the weight $\lambda_w$ of the wandering suppression penalty (WSP) to study its influence on navigation performance.
Specifically, we train the same model on Gibson while only varying $\lambda_w \in \{1.0,0.8,0.6,0.5,0.4,0.3,0.2,0.1\}$, and report SR and SPL on the Gibson validation set under the standard evaluation protocol.
All other training configurations (network architecture, RL hyperparameters, and data augmentations) are kept identical to ensure a controlled comparison.

\paragraph{Analysis.}
As shown in Table~\ref{tab:ablation_reward_weight}, $\lambda_w$ exhibits a non-monotonic effect on performance.
A large penalty (e.g., $\lambda_w=1.0$) overly suppresses revisits/backtracking, which can prevent necessary corrective behaviors and results in a clear drop (71.83\% SR / 65.86\% SPL).
As $\lambda_w$ decreases, both SR and SPL improve substantially, indicating that moderate suppression effectively reduces aimless wandering while preserving flexibility.
Notably, $\lambda_w=0.2$ achieves the \textbf{highest SR} of \textbf{94.00\%} while maintaining competitive SPL (71.20\%), reflecting the best overall robustness in goal-reaching.
Although $\lambda_w=0.5$ yields the highest SPL (73.30\%), it is accompanied by a lower SR (89.40\%), suggesting that a stronger penalty may occasionally hinder recovery from suboptimal states.
Therefore, we set $\lambda_w=0.2$ as the default in all experiments to prioritize consistent success in unseen environments while still keeping path efficiency at a strong level.

\subsection{Efficiency Analysis}
\label{app:efficiency}

To assess the practical deployment cost of HRNav, we evaluate its inference efficiency under the same setting as our real-world experiments. Specifically, we measure the average latency per step over 5,000 navigation steps on a server with one NVIDIA H20 GPU while varying the planning interval $k$ of the high-level planner. Here, the \emph{slow latency} denotes the runtime of one VLM-based planning call, the \emph{fast latency} denotes the runtime of one low-level policy step, and the \emph{average latency per step} amortizes the sparse slow-planner calls over the entire navigation process.

\paragraph{Analysis.}
Table~\ref{tab:efficiency} shows that HRNav maintains practical real-time performance despite introducing a VLM-based slow planner. With our default setting of $k{=}15$, the system runs at 24.29 FPS with an average latency of 41.16\,ms per step, which is sufficient for real-time robotic control. Although each slow-planner call takes around 370--400\,ms, the planner is only invoked sparsely, so its overhead is effectively amortized across subsequent execution steps. In contrast, the low-level executor remains consistently lightweight, requiring only 14--16\,ms per step across all settings.

The table also reveals a clear efficiency--performance trade-off. Using a smaller planning interval (e.g., $k{=}5$) provides slightly stronger navigation performance, but substantially increases average latency because the slow planner is called more frequently. Increasing the interval to 30 or 60 further improves throughput, but leads to a mild drop in SR and SPL due to less frequent planning updates. We therefore choose $k{=}15$ as the default setting, since it provides the best balance between navigation performance and computational efficiency.

\subsection{More visualization}

\begin{figure*}[t]
    \centering
    \includegraphics[width=0.95\linewidth]{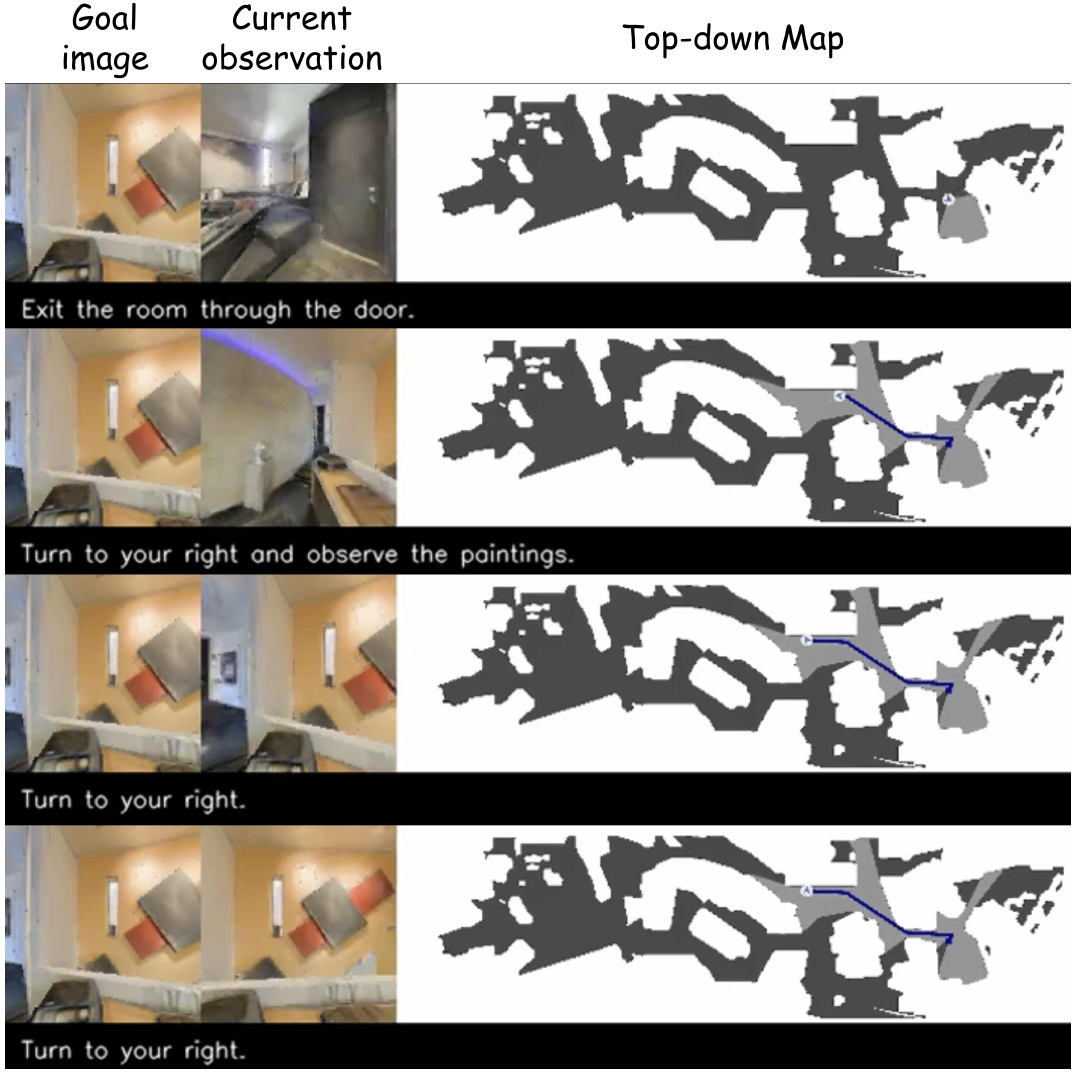}
    \caption{\textbf{Simulation qualitative result (top-down view).}
    Visualization of an example episode in simulation.
    The agent trajectory is shown in a global top-down map, illustrating the navigation progress toward the goal and the overall path efficiency.}
    \label{fig:vis_sim1}
\end{figure*}

\begin{figure*}[t]
    \centering
    \includegraphics[width=0.95\linewidth]{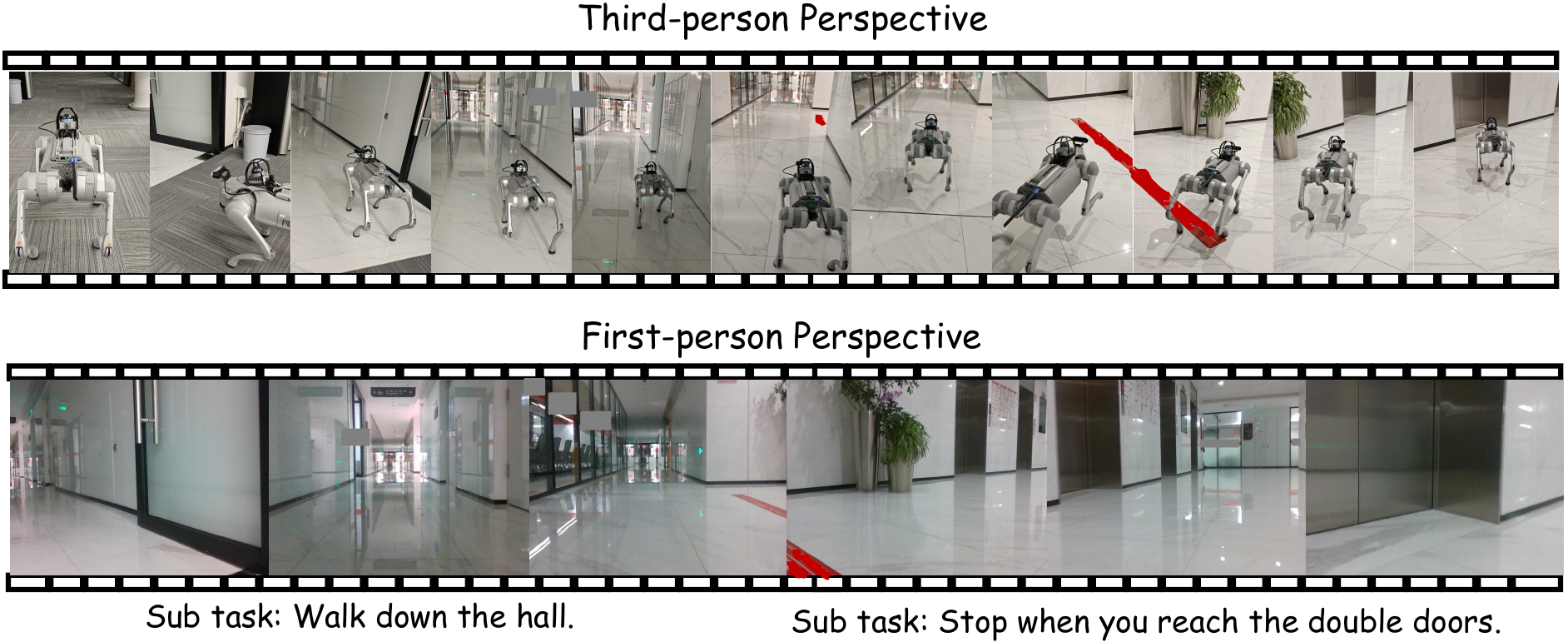}
    \caption{\textbf{Real-world qualitative result.}
    Visualization of an example episode in a real environment.
    We show third-person recordings and the robot’s egocentric observations along the timeline, together with the predicted sub-task sequence, to illustrate subtask-conditioned execution and successful goal reaching.}
    \label{fig:vis_real_world}
\end{figure*}

\paragraph{Qualitative visualizations.}
We provide additional qualitative results in both simulation and real-world environments to better illustrate the behavior of HRNav.
Fig.~\ref{fig:vis_sim1} shows a representative simulated episode in a top-down view, where we overlay the agent trajectory to reveal its global navigation pattern.
Fig.~\ref{fig:vis_real_world} presents a real-world episode with third-person and egocentric views along the timeline, highlighting how the robot progresses under the sub-task plan.

\paragraph{Analysis.}
Across both settings, HRNav exhibits more goal-directed trajectories with fewer redundant backtracks, indicating effective wandering suppression and stable long-horizon execution.
In simulation, the top-down view clearly shows that the agent follows a smoother and more direct route toward the goal, rather than oscillating between nearby locations.
In the real world, HRNav maintains consistent progress by executing actions conditioned on the current sub-task, and switches sub-tasks smoothly as the scene evolves.
These visualizations support our quantitative results, suggesting that hierarchical reasoning with subtask-conditioned control improves robustness, particularly in challenging long-horizon navigation.

\end{document}